\documentclass[10pt,twocolumn,letterpaper]{article}

\newcommand{\Tref}[1]{Table~\ref{#1}}
\newcommand{\Eref}[1]{Eq.~(\ref{#1})}
\newcommand{\Fref}[1]{Fig.~\ref{#1}}
\newcommand{\Sref}[1]{Sec.~\ref{#1}}

\renewcommand{\paragraph}[1]{\vspace{1mm}\noindent\textbf{#1}}
\newcommand{\bestcolor}{FFFC9E}

\PassOptionsToPackage{table,dvipsnames}{xcolor}
\usepackage[pagenumbers]{iccv} 
\usepackage{multirow}
\usepackage{algorithm}
\usepackage{algpseudocode}
\usepackage{amsmath}
\usepackage{amsfonts}
\usepackage{makecell}

\definecolor{iccvblue}{rgb}{0.21,0.49,0.74}
\usepackage[pagebackref,breaklinks,colorlinks,allcolors=iccvblue]{hyperref}

\title{CCMNet: Leveraging Calibrated Color Correction Matrices\\for Cross-Camera Color Constancy}

\author{
  Dongyoung Kim$^{1}$ \;
  Mahmoud Afifi$^{2}$ \;
  Dongyun Kim$^{1}$ \;
  Michael S. Brown$^{2}$ \; 
  Seon Joo Kim$^{1}$ \\[0.8ex]
  $^1$Yonsei University \quad $^2$AI Center - Toronto, Samsung Electronics \\[0.3ex]
  \footnotesize \texttt{\{dongyoung.kim,dongyunkim,seonjookim\}@yonsei.ac.kr} \quad \texttt{\{m.afifi1,michael.b1\}@samsung.com}
}

\begin{document}

\twocolumn[{%
\renewcommand\twocolumn[1][]{#1}%
\maketitle
\vspace{-11mm}

\begin{center}
\footnotesize{Project page: \href{https://www.dykim.me/projects/ccmnet}{\textcolor{iccvblue}{https://www.dykim.me/projects/ccmnet}}}
\end{center}
\begin{center}
\includegraphics[width=0.98\textwidth]{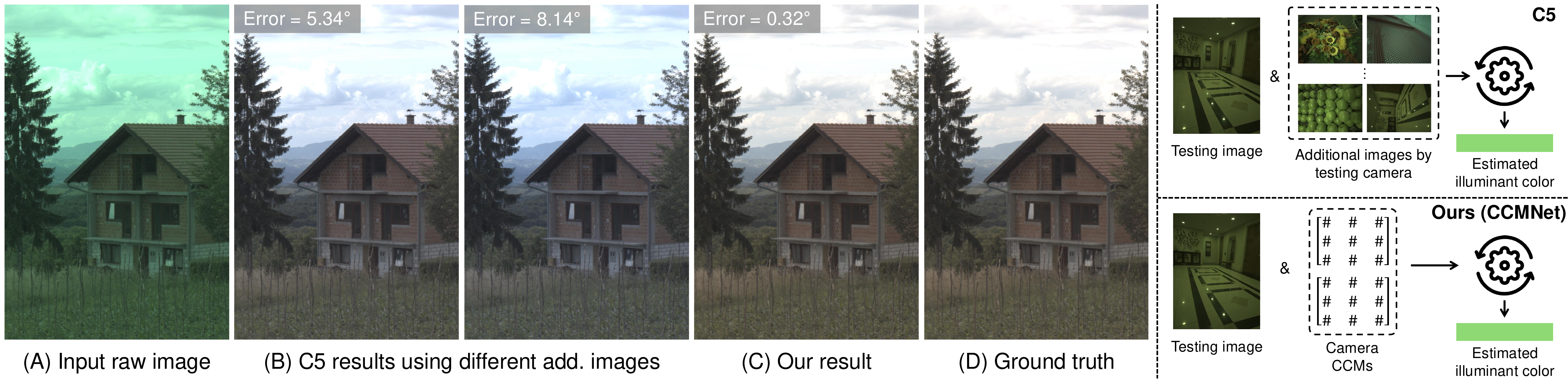}
\vspace{-2mm}
\captionof{figure}{This paper introduces CCMNet, a framework for cross-camera color constancy. CCMNet uses pre-calibrated color correction matrices (CCMs) from camera ISP hardware to train an encoder that generates a {\it camera fingerprint embedding} (CFE), capturing the testing camera’s color space. In (A), we show a raw image from a Canon 550D. In (B), we present C5 \cite{afifi2021cross}, which generalizes using randomly selected unlabeled images from the test camera---C5's performance varies depending on the image set. In (C), we show our results, relying only on fixed CCMs in the ISP. Neither method used Canon 550D data during training. Gamma correction was applied for visualization.} \label{fig:teaser}
\end{center}%
\vspace{-1mm}
}]

\begin{abstract}
Computational color constancy, or white balancing, is a key module in a camera's image signal processor (ISP) that corrects color casts from scene lighting. Because this operation occurs in the camera-specific raw color space, white balance algorithms must adapt to different cameras. This paper introduces a learning-based method for cross-camera color constancy that generalizes to new cameras without retraining.  Our method leverages pre-calibrated color correction matrices (CCMs) available on ISPs that map the camera's raw color space to a standard space (e.g., CIE XYZ). Our method uses these CCMs to transform predefined illumination colors (i.e., along the Planckian locus) into the test camera's raw space. The mapped illuminants are encoded into a compact camera fingerprint embedding (CFE) that enables the network to adapt to unseen cameras.  To prevent overfitting due to limited cameras and CCMs during training, we introduce a data augmentation technique that interpolates between cameras and their CCMs. Experimental results across multiple datasets and backbones show that our method achieves state-of-the-art cross-camera color constancy while remaining lightweight and relying only on data readily available in camera ISPs.
\vspace{-2mm}
\end{abstract}    
\section{Introduction}
\label{sec:intro}
Computational color constancy ensures that object colors remain consistent under varying lighting conditions \cite{barnard1995computational}. In digital cameras, this is achieved through white balancing, which adjusts raw image colors to simulate neutral lighting \cite{gijsenij2011computational, barnard2002comparison, cheng2015beyond}. This involves two main steps: illuminant estimation and linear white-balance correction \cite{delbracio2021mobile}.

Illuminant estimation predicts the color of the scene’s light source under the assumption of single-source illumination \cite{gijsenij2011computational}. The estimated illuminant color is then used in linear white-balance correction to counteract the effects of lighting and camera response biases \cite{florin2009color, brown2023color}. These steps are applied early in the image signal processor (ISP) pipeline to raw images \cite{afifi2019color} and are influenced by the camera's sensor-specific characteristics, such as response functions and lens properties \cite{nguyen2014raw, afifi2021semi}. These factors complicate the generalization of illuminant estimation algorithms across cameras with varying characteristics \cite{afifi2019sensor}.

Recent work on illuminant estimation achieves promising results using learning-based models \cite{gijsenij2010generalized, yu2020cascading, lo2021clcc, afifi2025optimizing}. These models learn a mapping between input image colors and scene illuminant colors using pairs of images and corresponding ground-truth illuminant colors, typically captured by the same camera used in testing \cite{oh2017approaching}. Consequently, most learning-based methods struggle to generalize to new cameras with different characteristics than those used during training \cite{afifi2019sensor}. This limitation hinders their practical applicability in manufacturing, as fine-tuning or retraining is necessary for each new camera introduced. Some recent attempts have proposed solutions for improved adaptation through few-shot learning \cite{xiao2020multi}, or by using additional unlabeled images captured by the testing camera at inference time to facilitate generalization to the testing camera’s color space \cite{afifi2021cross}. While these methods show promising results, they require capturing new images with the testing camera for adaptation \cite{li2023ranking}, making their performance inherently dependent on the characteristics of those images \cite{afifi2021cross}. 

While camera ISPs rely on pre-calibrated, camera-specific information to assist in color processing after white balance has been applied, to the best of our knowledge, no prior work has leveraged this calibrated information for the cross-camera color constancy task. Specifically, consumer-grade ISPs rely on calibrated color correction matrices (CCMs) to transform the camera’s raw color space to a device-independent standard color space (e.g., CIE XYZ). These CCMs are carefully calibrated during ISP manufacturing \cite{karaimer2018improving}, are easily accessible within the ISP's firmware \cite{hasinoff2016burst, delbracio2021mobile, brown2023color}, and are also available in DNG files for post-capture raw rendering \cite{adobe2023digital}. The availability of this information motivated us to utilize this calibrated data to improve cross-camera generalization.

\indent{{\bf Contribution}}~In this paper, we present CCMNet, a learning-based method for cross-camera color constancy built on the convolutional color constancy (CCC) framework~\cite{hubel2007white, barron2015convolutional, barron2017fast, afifi2021cross}. Our method leverages pre-calibrated color correction matrices (CCMs) available from camera ISPs to transform predefined illuminant colors along the Planckian locus from the device-independent CIE XYZ color space into the raw space of the test camera. These transformed illuminations encode the unique characteristics of the camera’s response function and serve as reference points. The transformed illuminations are compressed into an 8-dimensional embedding, allowing a learnable hypernetwork to adapt to the test camera’s raw color space and generate a camera-specific CCC model tailored to the input image.  Additionally, we introduce an augmentation technique that maps training images from a limited set of cameras to imaginary raw spaces, improving generalization. Consequently, CCMNet, which combines a design that dynamically adapts to the raw space of various cameras with a robust data augmentation strategy, accurately estimates illuminant colors for cameras unseen during training (see Fig. 1-C). Our approach is lightweight, accurate, and requires no additional test camera images, unlike prior work \cite{afifi2021cross}.
\section{Related Work}
\label{sec:related}
A camera's ISP includes several color processing modules applied in a pipeline fashion. One of the early-stage modules corrects the colors of the captured raw image through two key steps \cite{florin2009color, karaimer2016software, brown2023color}: (1) image white balancing (\Sref{sec:awb}) and (2) transferring the camera raw colors to a standard color space via CCMs (\Sref{sec:ccm}).

\subsection{Auto White Balance}
\label{sec:awb}
As discussed in \Sref{sec:intro}, auto white balance modules consist of two steps: illuminant estimation and correction. The correction is applied to the linear raw image colors, often using a diagonal correction matrix \cite{barnard1995computational}. Most research focuses on illuminant estimation, which determines the scene’s illuminant color in the camera’s raw space. This can be categorized into learning-free (statistical) methods (e.g., \cite{buchsbaum1980spatial, land1977retinex, finlayson2004shades, van2007edge, gijsenij2011improving, cheng2014illuminant, qian2018revisiting, qian2019finding, ulucan2024multi}) and learning-based methods (e.g., \cite{barron2015convolutional, hu2017fc4, lo2021clcc, tang2022transfer, afifi2025optimizing}).

Statistical-based methods rely on specific hypotheses (e.g., gray-world \cite{buchsbaum1980spatial}, gray-edges \cite{van2007edge}, etc.) and use statistics derived from the input raw image colors to estimate the illuminant color. As a result, they inherently generalize across different cameras. However, these methods often have limited accuracy and may fail in scenarios where the scene's illuminant color cannot be reliably inferred from the captured image.

Learning-based methods (e.g., \cite{gehler2008bayesian, gijsenij2010generalized, bianco2015color, barron2015convolutional, shi2016deep, hu2017fc4, oh2017approaching, barron2017fast, xu2020end, qian2017recurrent, yu2020cascading, lo2021clcc, yue2023color, kim2021large, kim2024attentive, li2024nightcc}) improve accuracy by training models to map raw colors to illuminant colors. However, most fail to generalize to unseen cameras \cite{afifi2019sensor}. Some approaches attempt adaptation via meta-learning and few-shot learning \cite{mcdonagh2018formulating}, assuming access to a range of illuminant colors from the testing camera \cite{hernandez2020multi, yue2024effective}, or creating generic methods that require fine-tuning on the new camera \cite{bianco2019quasi}.

Among these efforts, our work falls into a category of methods designed to achieve adaptation without requiring a paired set of images from the test camera, even if that set is small. To this end, the work in \cite{afifi2019sensor} (termed SIIE) maps input raw images from different cameras to a learnable {\it working space}, reducing disparities in raw color spaces before illuminant estimation. However, this method assumes access to a diverse range of training cameras to effectively learn this mapping, making its accuracy dependent on the variability of the training data. More recently, C5 \cite{afifi2021cross} utilizes additional images captured by the test camera during inference to dynamically generate a CCC model \cite{barron2015convolutional, barron2017fast}. While this method achieves promising results, its accuracy heavily depends on the characteristics of the additional images provided (see \Fref{fig:teaser}-B).

In contrast, our method leverages a static set of predefined guidance colors along with pre-calibrated data from the test camera, enabling consistently high accuracy without requiring additional images from the test camera.

\subsection{Color Space Transfer via CCMs}
\label{sec:ccm}
Camera sensors exhibit unique spectral sensitivity and bias, resulting in each camera having its own native RGB color space. Camera ISP manufacturers calibrate and apply color correction matrices (CCMs) to facilitate image processing, ensuring an appropriate transformation between the native RGB space and a device-independent standard color space (e.g., CIE XYZ) within the imaging pipeline~\cite{bianco2013color, karaimer2016software}.

\begin{figure}
\centering
\includegraphics[width=1.0\linewidth]{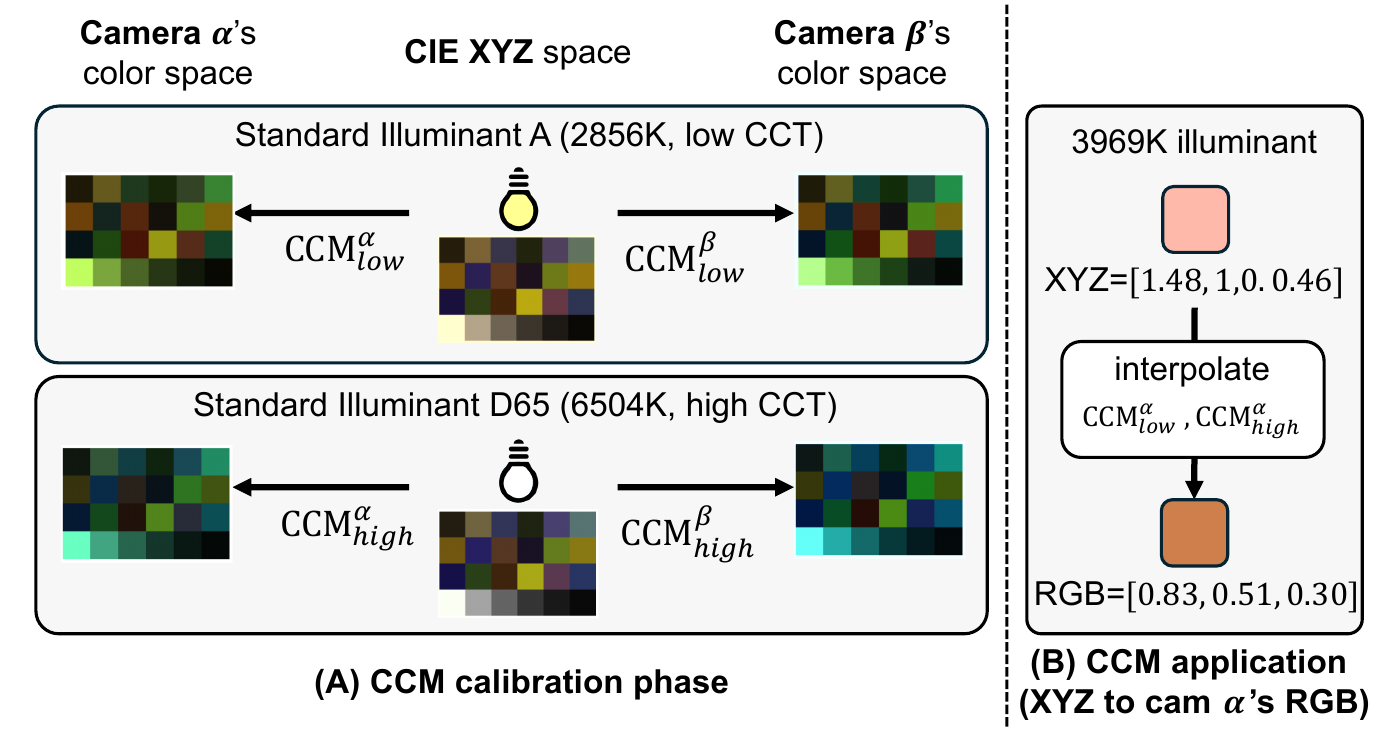}
\vspace{-6mm}
\caption{Example of CCM calibration (A) and application (B). CCMs are calibrated to transform between CIE XYZ chromaticity and camera-specific raw-RGB values under standard illuminants with predefined color temperatures (e.g., 2856K, 6504K). For other illuminants, the calibrated CCMs are interpolated. As a result, CCMs reflect the camera's unique color characteristics, capturing how the camera \textit{perceives} illuminants along the color temperature trajectory.}
\vspace{-4mm}
\label{fig:related-work}
\end{figure}

Although the transformation between the camera’s raw space and a standard color space is often nonlinear~\cite{finlayson2015color, finlayson2020designing, hong2001study, hung1993colorimetric}, cameras primarily rely on linear transformation matrices due to their simplicity and practical benefits~\cite{finlayson2020designing}. CCMs are typically calibrated by fitting a  $3 \times 3$ matrix that maps the raw RGB values of a calibration object (e.g., a color chart) to their corresponding values in a standard color space under an illuminant with a specific correlated color temperature (CCT)~\cite{karaimer2018improving, adobe2023digital}. To accommodate diverse lighting conditions, CCMs are precomputed for at least two illuminants (typically for low and high CCTs~\cite{mcelvain2013camera}) and interpolated for intermediate conditions (see \Fref{fig:related-work}).

CCMs serve as the critical link between a camera's unique color characteristics and a standard color space. While most CCMs are designed to transform white-balanced camera raw-RGB to CIE XYZ, some types of CCMs within the ISP operate in the reverse direction, mapping observed CIE XYZ under a specific illuminant back to the camera’s native raw-RGB space, as shown in \Fref{fig:related-work}-A. This inverse transformation, in particular, provides insight into how various illuminants are represented in the native raw-RGB space. By leveraging this transformation, we can approximate the color trajectories of illuminants in the raw-RGB domain across a range of CCTs.

We leverage this property of CCMs as a {\it bridge} to introduce a novel illuminant estimation method that adapts to the color space of unseen cameras. While previous studies have leveraged CCMs for data augmentation~\cite{afifi2021cross},  none have explored their use during inference to improve illuminant estimation across different cameras. Additionally, we introduce a data augmentation strategy that exploits the linearity of CCMs to enhance generalization. Specifically, we propose a technique to generate {\it imaginary} camera images with corresponding CCMs, further improving the robustness and adaptability of our model.

\section{Method}
\label{sec:method}

\begin{figure*}[ht]
\includegraphics[width=\textwidth]{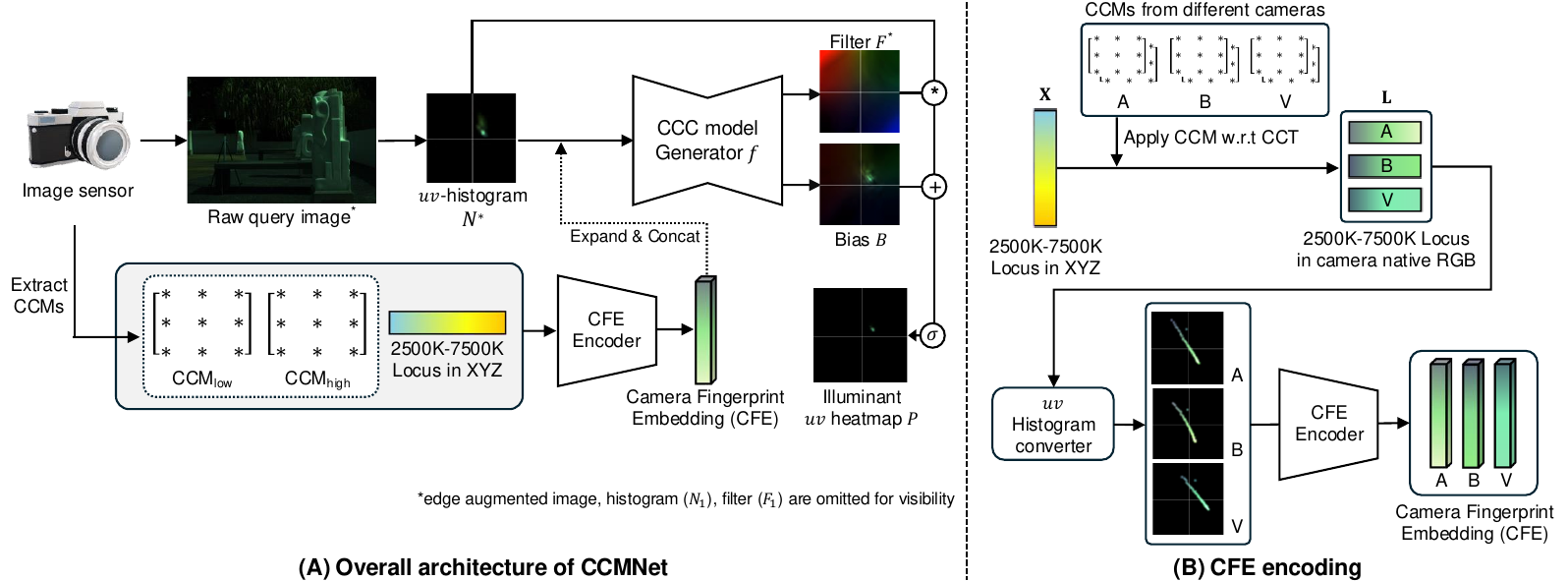}
\caption{Overview of the CCMNet architecture. (A) Based on CCC \cite{barron2015convolutional} and C5 \cite{afifi2021cross}, CCMNet includes a network $f$ that generates filters and bias from the $uv$-histograms of the input image. To process query images from diverse camera domains with varying spectral sensitivities, CCMNet uses a camera fingerprint embedding (CFE) as guidance. (B) The CFE for three example cameras (A, B, V)---two real (A, B) and one imaginary (V)---is constructed by mapping predefined illuminants (2500K–7500K along the Planckian locus) from the CIE XYZ space to each camera's native raw RGB space using calibrated CCMs. These values are converted into a $64 \times 64$ histogram and encoded into a 1D vector via a lightweight encoder.}
\vspace{-4mm}
\label{fig:model_architecture}
\end{figure*}

\subsection{Preliminary}
\label{sec:overview}
\paragraph{Auto White Balance Formulation.}
Assuming a single global illumination, a given linear raw image, $I$, is formed as the element-wise product of its white-balanced counterpart, $W$, and the global illuminant RGB color vector, $\ell$, at every pixel location $x$. This can be mathematically described as follows:
\begin{equation}
     I^{x} = W^{x} \circ \ell \quad  \forall_{x}.
\end{equation}

The conventional goal of the auto white-balance task is to optimize a model  $f$ that estimates the illumination RGB from a given raw image $I$:
\begin{equation}
    [\hat{\ell}_R, \hat{\ell}_G, \hat{\ell}_B]^T = f(I).
\end{equation}

\paragraph{Convolutional Color Constancy (CCC).}
As shown in the upper flow of \Fref{fig:model_architecture}-A, our method is fundamentally based on the CCC framework \cite{barron2015convolutional, barron2017fast}, which transforms the image histogram $N$ into an illuminant heatmap $P$, using a filter $F$ and a bias $B$.
CCC reformulates the illumination estimation problem as a coordinate localization task on a log-chroma histogram \cite{finlayson2001color}, commonly termed a $uv$-histogram.
Specifically, for an image’s RGB pixel $[I_R,I_G,I_B]$, the log-chroma values, $u$ and $v$, are calculated as follows (pixel coordinate $x$ is omitted for simplicity):
\begin{equation}
    I_u = \log(I_G /I_R),\; I_v = \log(I_G/I_B).
\label{eq3:uv_conversion}
\end{equation}

After that, a $uv$-histogram can be generated as follows:
\begin{equation}
    N(u,v) = \sum_{x} \left\| I^x \right\|_2 [\left| I_u^x - u\right|\leq \epsilon \, \wedge \left| I_v^x - v\right|\leq \epsilon],
\label{eq4:hist_conversion}
\end{equation}
\noindent where $\epsilon$ is the width of the histogram bin and $\left\| I^x \right\|_2$ is the weighting factor for each pixel, defined as the L2 norm of the raw RGB values of the pixel. In other words, the value of  $N(u,v)$ represents the weighted count of pixels in image $I$ that fall within a certain range ($\epsilon$) around the point $(u,v)$.

The goal of CCC is to optimize a global filter $F$ and bias $B$, to predict a probability map $P$  of the illumination within the histogram space using the following equation:
\begin{equation}
    P = \sigma(B + \sum_{i}(N_i * F_i)),
\label{eq5:uv_prediction}
\end{equation}
\noindent where $F$ and $B$ have the same shape as the $uv$-histogram N, $*$ represents the convolution operation (accelerated using fast Fourier transforms), $\sigma$ represents the softmax operation over the $uv$-coordinate space, and the subscript $i$ denotes the index corresponding to the filter and histogram. Here, $i=0$ refers to the original raw image, while $i \geq 1$ corresponds to augmented images (e.g., texture, edge).
We can interpret $P$ as a heatmap of confidence for each $u,v$ coordinate, so the final prediction $(\hat{\ell}_u, \hat{\ell}_v)$ is expressed as a weighted sum of the coordinates using $P$:
\begin{equation}
    \hat{\ell}_u  = \sum_{u,v} u P(u, v), \;\;\;
    \hat{\ell}_v  = \sum_{u,v} v P(u, v).
\label{eq6:weighted_uv}
\end{equation}
The final RGB illumination estimate, $\hat{\ell}$, is obtained by inverting the transformation in \Eref{eq3:uv_conversion}:
\begin{equation}
    \hat{\ell} = \left[ \exp\left(-\hat{\ell}_u\right), 1, \exp\left(-\hat{\ell}_v\right) \right],
\label{eq7:final_rgb}
\end{equation}
where the green channel is assumed to be G=1.
Alternatively, $\hat{\ell}$ can be normalized to ensure that the vector has unit length.

The training objective of CCC is to optimize the filter and bias to minimize the angular error between the predicted illumination RGB and the ground truth illumination in the training dataset. For cross-camera color constancy, C5 \cite{afifi2021cross} proposes a hypernetwork version of CCC that dynamically generates $F$ and $B$ for the test image after analyzing histograms of additional images taken by the same camera.

\subsection{CCMNet}
The proposed CCMNet framework is built upon C5 \cite{afifi2021cross}. As mentioned above, C5 is a hypernetwork version of CCC \cite{barron2015convolutional, barron2017fast}, where the network dynamically generates filters and bias.
However, unlike C5, CCMNet does not require additional images from the target camera (typically 6–8). Instead, our method leverages two pre-calibrated Color Correction Matrices (CCMs) for low and high correlated color temperatures (CCTs) embedded within the ISP (see \Fref{fig:model_architecture}-A). These CCMs provide stable guidance via Camera Fingerprint Embedding (CFE) (see \Fref{fig:model_architecture}-B), ensuring consistent performance across diverse camera domains without extra test images. The core formulation of CCMNet is as follows:
{\small
\begin{equation}
\left\{ F_0, F_1, B \right\} = \textrm{CCMNet}(N_0, N_1, \textrm{CCM}_{low}, \textrm{CCM}_{high}),
\label{eq8:formulation}
\end{equation}
}

\noindent where $N_0$ and $N_1$ denote the original raw image and its edge-augmented counterpart, $\textrm{CCM}_{low}$ and $\textrm{CCM}_{high}$ are pre-calibrated matrices corresponding to low and high CCTs (typically 2500K and 6500K). The outputs $F_0$, $F_1$, and $B$, generated by CCMNet, are used to estimate the final $uv$ coordinate of the illumination through Eqs. \eqref{eq5:uv_prediction}--\eqref{eq7:final_rgb}.

In \Sref{sec:cfe}, we introduce Camera Fingerprint Embedding (CFE), a method for extracting device-specific guidance features using CCMs. Additionally, in \Sref{sec:augmentation}, we propose an imaginary camera augmentation technique to mitigate overfitting to the limited number of cameras and CCMs used during training. These strategies (CFE and augmentation) enhance CCMNet’s ability to generalize across diverse spectral sensitivities, achieving state-of-the-art performance in cross-camera color constancy tasks.

\subsection{Camera Fingerprint Embedding}
\label{sec:cfe}
Cross-camera color constancy aims to estimate the chromaticity of the light source while adapting to unseen sensor domains. To this end, we introduce a device-aware guidance feature called Camera Fingerprint Embedding (CFE). CFE encodes the color trajectory of light sources \textit{observed} by each camera within a specific color temperature range into an 8-dimensional vector. As a result, it inherently captures each camera’s unique color characteristics, enabling the model to adapt to the color space of previously unseen cameras. As shown in \Fref{fig:model_architecture}-B, CFE is generated through a two-step process. First, a set of illuminants along the Planckian locus (covering color temperatures from 2500K to 7500K) is converted into the specific camera’s native raw RGB space using its pre-calibrated CCMs. Second, the resulting RGB illuminant colors are transformed into $uv$-histogram, which is then processed by a CNN-based encoder to extract device-specific feature.

\paragraph{Camera-Native Guidance Illuminants.}
Our goal is to obtain ${\mathbf{L}}$, the chromaticity set of light sources within a specific color temperature range as observed by a given camera. To achieve this, we use calibration matrices that transform the CIE XYZ coordinates of standard illuminants A or D65 into the camera’s raw space. These calibration data are typically provided during camera manufacturing and can be extracted from DNG files produced by most cameras. For simplicity, we refer to these matrices as $\textrm{CCM}_{low}$ and $\textrm{CCM}_{high}$ throughout this paper, as used in \Eref{eq8:formulation}. For details on CCM properties and extraction methods, please refer to the supplementary materials.

First, illuminant colors along the Planckian locus in the device-independent CIE XYZ color space are sampled at 100K intervals within the 2500K to 7500K color temperature range. Each sampled XYZ point, $\mathbf{X}_t$, corresponding to an illuminant with color temperature $t$, is then transformed into a camera-native RGB color, $\mathbf{L}_t$, using the following equation:
\begin{equation}
\mathbf{L}_t = \textrm{CCM}_t \mathbf{X}_t,
\label{eq9:xyz2raw}    
\end{equation}
where $\textrm{CCM}_t$ is a transformation matrix for color temperature $t$ that maps the CIE XYZ values of an illuminant with color temperature $t$ to the target camera’s raw space. Since $\textrm{CCM}_{low}$ and $\textrm{CCM}_{high}$ are calibrated at specific color temperatures, interpolation is used to compute $\textrm{CCM}_t$ for an arbitrary color temperature $t$. The interpolation of $\textrm{CCM}_t$ is defined as:
\begin{equation}
\begin{aligned}
\textrm{CCM}_t &= g \textrm{CCM}_{low} + (1-g) \textrm{CCM}_{high}, \\\\
\textrm{where} \;g &= \frac{t^{-1}-\textrm{CCT}_{high}^{-1}}{\textrm{CCT}_{low}^{-1}-\textrm{CCT}_{high}^{-1}},
\label{eq10:ccm_interp}
\end{aligned}
\end{equation}
\noindent where $\textrm{CCT}_{low}$ and $\textrm{CCT}_{high}$ denote the color temperatures of the standard illuminants for which $\textrm{CCM}_{low}$ and $\textrm{CCM}_{high}$ are calibrated, typically around 2500K and 6500K, respectively.
The resulting set of camera-native RGB colors, ${\mathbf{L}_t \mid t \in \{2500, 2600, \dots, 7500\}}$, represents the illumination colors along the Planckian locus in the camera’s raw RGB space, sampled within the 2500K–-7500K range as observed by a specific image sensor.

\paragraph{Histogram Conversion \& Encoding.} The camera-specific guidance illuminant set, ${\mathbf{L}}$, is transformed into a $uv$-histogram using \Eref{eq3:uv_conversion} and \Eref{eq4:hist_conversion}. As shown in \Fref{fig:model_architecture}-B, the guidance illumination set follows distinct trajectories for each device in the $uv$-histogram space. To convert these trajectory differences into a device-aware embedding, we employ a lightweight CNN, the CFE encoder, consisting of four convolutional layers (with max pooling) followed by a two-layer MLP. This network encodes each device’s locus histogram into an 8-dimensional CFE feature. The encoded CFE feature is then repeated along the $u$ and $v$ axes to match the resolution of the input histograms. Finally, it is concatenated with the input histograms ($N_0$, $N_1$) along the channel dimension and provided as input to the CCC generator network $f$, as shown in \Fref{fig:model_architecture}-A.

\subsection{Imaginary Camera Augmentation}
\label{sec:augmentation}
In prior illuminant estimation research, most data augmentation techniques (e.g., \cite{fourure2016mixed, lou2015color, abdelhamed2021leveraging, Afifi2020CIEXN}) rely on transferring ground-truth illuminant colors---randomly sampled from a given dataset---to other images within the same dataset (captured by the same camera) using chromatic adaptation. However, this approach is incompatible with our method, which is trained on raw images from different cameras, each with a distinct raw color space. Another augmentation approach \cite{afifi2021cross} leverages camera-specific information and CCMs to perform raw-to-raw augmentation by transferring images from a source camera to a target camera. While promising, this method remains constrained by the limited diversity of training camera raw spaces.

To address these limitations, we propose a novel augmentation strategy that increases the diversity of camera characteristics, even with a limited set of training cameras. Specifically, we synthesize \textit{imaginary} cameras by leveraging the CCMs of available training cameras. This expands the range of camera raw spaces encountered during training, significantly enhancing generalization.

\paragraph{Imaginary Camera Image Synthesis.} Under the assumption of a single illuminant in the scene, the value of channel $c\in \{R, G, B\}$ at pixel $x$ in the camera's raw space can be expressed as:
\begin{equation}
I_c(x) = \int S(\lambda, x) R(\lambda) Q_c(\lambda) \, d\lambda,
\end{equation}
where $S(\cdot)$ and $R(\cdot)$ represent the spectral power distribution of the scene and illuminant at pixel $x$, respectively, and $Q_c$ is the camera’s spectral sensitivity for color channel $c$. The integral is computed over $\lambda$, corresponding to wavelengths in the visible light spectrum.

\begin{figure}[t]
\centering
\includegraphics[width=1.0\linewidth]{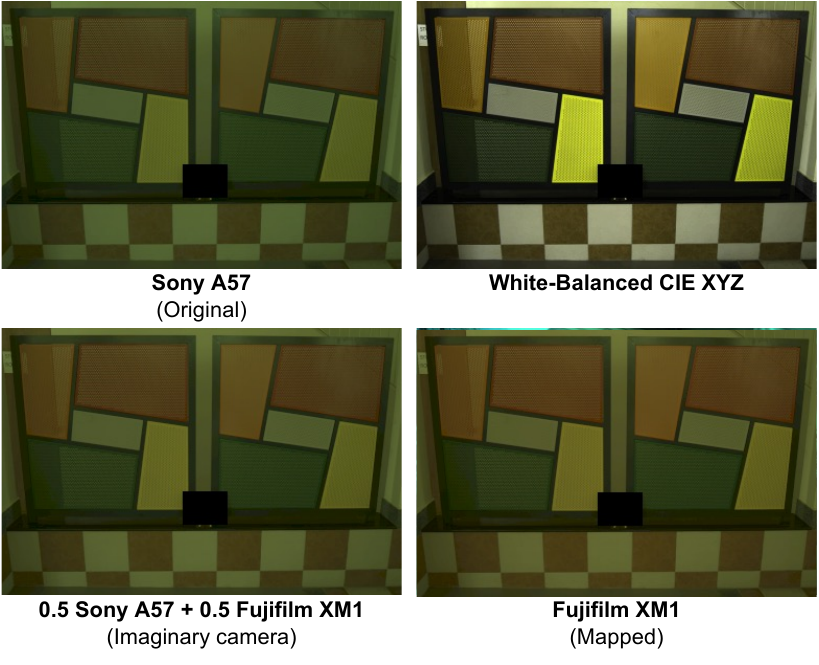}
\vspace{-5mm}
\caption{Visualization of our imaginary camera augmentation process. An image from the Sony A57 is white-balanced using the ground-truth illuminant, converted to CIE XYZ space, and mapped to the target camera’s raw space. We illustrate two cases: mapping to the raw space of a real camera (Fujifilm XM1) and an imaginary camera. Brightness is adjusted for clarity.}
\label{fig:imagin_cam}
\end{figure}

Since a camera’s characteristics are defined by its spectral sensitivity function $Q$, an image captured by an \textit{imaginary} camera, denoted as $V$, can be approximated by linearly combining the characteristics of cameras $A$ and $B$ with a ratio $\alpha$, defined as:
\begin{equation}
\begin{aligned}
I_{c}^V &= \int S(\lambda) R(\lambda) ( \alpha Q_{c}^A + (1 - \alpha) Q_{c}^B)(\lambda) \, d\lambda
\\  &= \alpha I_{c}^A + (1-\alpha)I_{c}^B \;,
\label{eq12:virtual_image}
\end{aligned}
\end{equation}
where superscripts $A$, $B$, and $V$ denote different cameras, including the imaginary camera, and $\alpha \in [0, 1]$ controls the contribution of each camera to the synthesized imaginary camera (omitting $x$ for simplicity). As illustrated in \Fref{fig:imagin_cam}, this approach allows mapping raw images to a specific target camera (e.g., Sony A57 with $\alpha = 1$) or to an imaginary camera (e.g., blending the Sony A57 and Fujifilm XM1 with a 0.5 to 0.5 ratio). Additionally, the ground truth illumination for the imaginary camera $V$ can be synthesized as a linear combination of the ground truth illuminations of cameras $A$ and $B$, weighted by $\alpha$. For additional details on augmentation methods, please refer to the supplementary materials.

\paragraph{Derivation of the Imaginary Camera’s CCM.}
Since CCMNet requires CCMs to encode CFE, it is also necessary to derive CCMs for the imaginary camera. Let us assume that cameras $A$, $B$, and the imaginary camera $V$ observe a light source with a correlated color temperature $\text{CCT}_{\text{low}}$. Based on \Eref{eq9:xyz2raw} and \Eref{eq12:virtual_image}, the observed raw RGB values for camera $V$ can be obtained as follows:

\begin{equation}
\begin{aligned}
\small
\textrm{CCM}_{low}^V \mathbf{X} &= \mathbf{L}^{V} = \alpha \mathbf{L}^{A} + (1 - \alpha) \mathbf{L}^{B} \\
&= \alpha (\textrm{CCM}_{low}^A  \mathbf{X}) + (1 - \alpha) (\textrm{CCM}_{low}^B \mathbf{X}) \\
&= [ \alpha \textrm{CCM}_{low}^A + (1 - \alpha) \textrm{CCM}_{low}^B ] \mathbf{X},
\end{aligned}
\end{equation}

\noindent where the superscript $V,A,B$ denotes the type of camera (omitted the subscript $\text{CCT}_{\text{low}}$ for $\mathbf{L}$ and $\mathbf{X}$ for simplicity). As a result, the $\textrm{CCM}_{low}$ for the imaginary camera $V$ can be defined as:
\begin{equation}
\textrm{CCM}_{low}^V = \alpha \textrm{CCM}_{low}^A + (1 - \alpha) \textrm{CCM}_{low}^B.
\end{equation}

This relationship also holds for $\textrm{CCT}_{high}$ and any arbitrary color temperature $t$ within the range of low and high CCTs in the calibrated CCMs, as described in \Eref{eq10:ccm_interp}. We randomly select two cameras from the training dataset to generate an augmented set using the method outlined above. The augmented images and CCMs approximate the spectral sensitivity of the imaginary camera, enabling the CFE encoder to generalize to a wider range of cameras despite the limited number of training cameras.
\section{Experiments}

\subsection{Experimental Setup}
\label{sec:setup}
\paragraph{Training.}
The input image and camera-specific raw RGB illuminants (51 colors ranging from 2500K to 7500K, sampled at 100K intervals) are represented as $64 \times 64$ $uv$-histograms. The $uv$-ranges are empirically set to [-2.85, 2.85] for the input query image and [-0.5, 1.5] for CFE encoding.

We use the Intel-TAU \cite{laakom2021intel}, Gehler-Shi \cite{shi2000re}, NUS-8 \cite{cheng2014illuminant}, and Cube+ \cite{banic2017unsupervised} datasets for training and testing. Each dataset includes images captured by distinct cameras, with no overlap between datasets. The number of cameras varies between one (Cube+) and eight (NUS-8).

Following the protocol in C5 \cite{afifi2021cross}, we adopt a leave-one-out cross-dataset evaluation approach, where the network is trained on all datasets except the test dataset. For instance, when validating on Gehler-Shi, the network is trained using Intel-TAU, NUS-8, and Cube+, ensuring no camera overlap between training and test datasets. We exclude the Sony-IMX subset of Intel-TAU due to the absence of CCM information, so Intel-TAU is used solely for training.

The mean angular error serves as the loss function during training. Additional details on batch size, epochs, other training hyperparameters, and model architecture are provided in the supplementary materials.

\paragraph{Data Augmentation.}
We augment the training data by selecting two cameras from the training datasets and applying camera-to-camera mapping with random ratio interpolation to generate images and CCMs for imaginary cameras, as described in \Sref{sec:augmentation}. The total number of augmented images matches the size of the original training set. Further details are provided in the supplementary materials.

\paragraph{Testing.}
For evaluation, we report commonly used error statistics: the mean, median, and tri-mean angular errors, along with the arithmetic mean of the top and bottom 25\% angular errors between the predicted and ground truth illuminations.

\subsection{Results}
\label{sec:results}

\begin{figure*}[t]
\centering
\includegraphics[width=\textwidth]{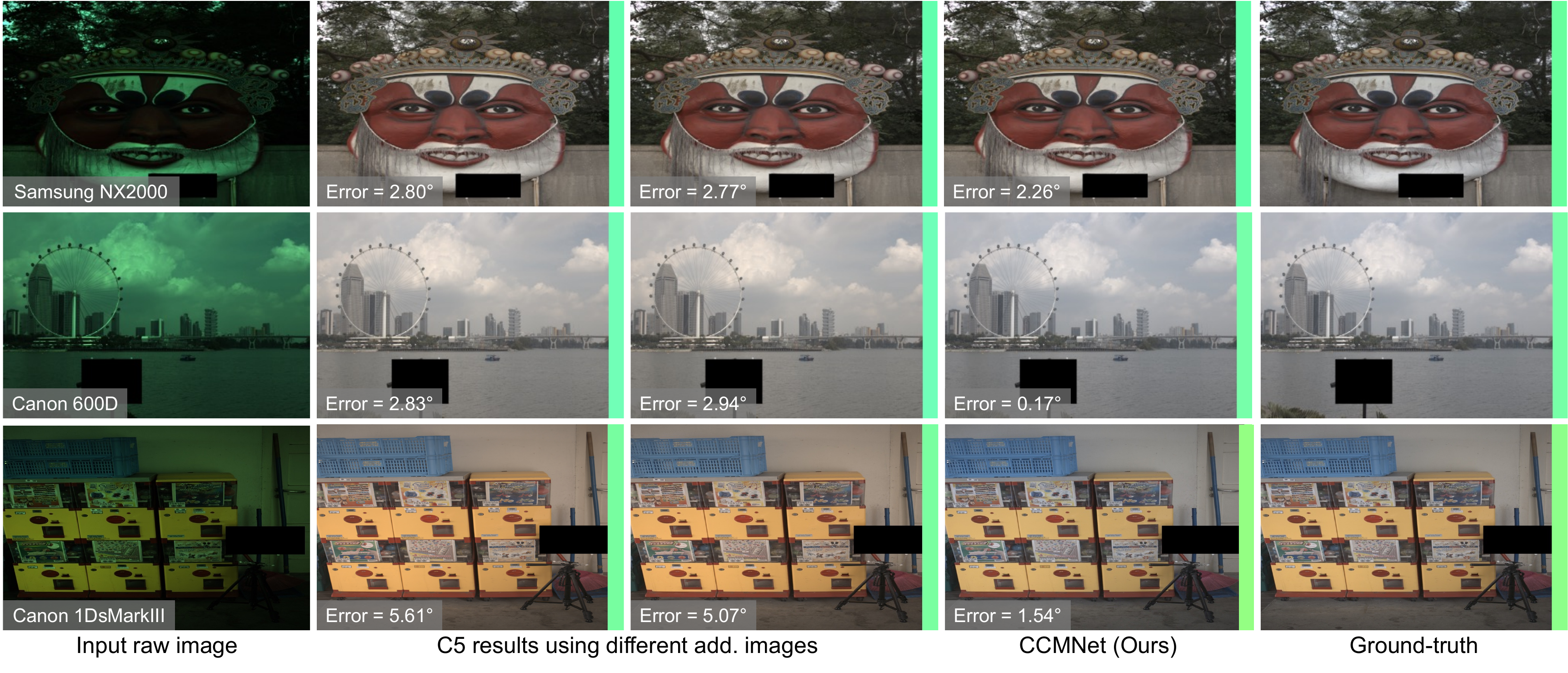}
\vspace{-6mm}
\caption{Visual comparison of the results from C5 \cite{afifi2021cross} with different additional image sets (second, third column) and CCMNet (fourth column). While C5 relies on additional images, CCMNet is optimized for fixed CFE guidance, ensuring consistent performance.}
\label{fig:quali_results}
\end{figure*}

Results are presented in \Tref{tab:main_table}, where the first three tables show the main experimental results for three test datasets: Cube+, Gehler-Shi, and NUS-8. These results demonstrate that CCMNet achieves state-of-the-art performance across all datasets and metrics (see \Fref{fig:quali_results}).

Unlike other learning-based models that report zero-shot results, DMCC retrains a target camera-specific network using calibrated matrices to transform training data (the Sony IMX-135 subset from the Intel-TAU dataset) into the test camera’s color space. Similarly, C5 requires additional images from the test camera for guidance, making it difficult to determine the optimal number and content of these images.

\begin{table}
\centering

\resizebox{0.95\linewidth}{!}{
\begin{tabular}{lccccc|c}
\toprule
\textbf{Gehler-Shi \cite{shi2000re}} & \textbf{Mean} & \textbf{Med.} & \textbf{Tri.} & \textbf{B.25\%} & \textbf{W.25\%} & \textbf{Size(MB)} \\ \midrule \midrule
2nd-order Gray-Edge \cite{van2007edge} & 5.13 & 4.44 & 4.62 & 2.11 & 9.26 & -  \\
Shades-of-Gray \cite{finlayson2004shades}& 4.93 & 4.01 & 4.23 & 1.14 & 10.20 & -   \\
PCA-based B/W Colors \cite{cheng2014illuminant}& 3.52 & 2.14 & 2.47 & 0.50 & 8.74 & -  \\
ASM \cite{Akbarinia2018ColourCB} & 3.80 & 2.40 & 2.70 & - & -  & - \\
Woo \textit{et al.} \cite{woo2017improving} & 4.30 & 2.86 & 3.31 & 0.71 & 10.14 & - \\
Grayness Index \cite{qian2019finding} & 3.07 & 1.87 & 2.16 & 0.43 &  7.62 & -  \\
Cross-dataset CC \cite{koskinen122020cross} & 2.87 & 2.21 & - & - & -  & - \\
Quasi-Unsupervised CC \cite{bianco2019quasi} & 3.46 & 2.23 & - & - & - & 622  \\
SIIE \cite{afifi2019sensor} & 2.77 & 1.93 & - & 0.55 & 6.53 & 10.3 \\
FFCC \cite{barron2017fast} & 2.95 & 2.19 & 2.35 & 0.57 & 6.75 & 0.22 \\
C5 ($m=7$) \cite{afifi2021cross} & 2.36 & 1.61 & 1.74 & 0.44 & 5.60 & 1.74 \\
C5 ($m=9$) \cite{afifi2021cross} & 2.50 & 1.99 & 2.03 & 0.53 & 5.46 & 2.09 \\
\textbf{CCMNet (Ours)} & \cellcolor[HTML]{\bestcolor}\textbf{2.23} & \cellcolor[HTML]{\bestcolor}\textbf{1.53} & \cellcolor[HTML]{\bestcolor}\textbf{1.62} & \cellcolor[HTML]{\bestcolor}\textbf{0.36} & \cellcolor[HTML]{\bestcolor}\textbf{5.46} & 1.05 \\ 
\bottomrule \\
\end{tabular}}
\vspace{-2mm}

\resizebox{0.95\linewidth}{!}{
\begin{tabular}{lccccc}
\toprule
\textbf{Cube+ \cite{banic2017unsupervised}} & \textbf{Mean} & \textbf{Med.} & \textbf{Tri.} & \textbf{B.25\%} & \textbf{W.25\%} \\ \midrule \midrule
Gray-world \cite{buchsbaum1980spatial} & 3.52 & 2.55 & 2.82 & 0.60 & 7.98 \\
1st-order Gray-Edge \cite{van2007edge} & 3.06 & 2.05 & 2.32 & 0.55 & 7.22 \\
2nd-order Gray-Edge \cite{van2007edge} & 3.28 & 2.34 & 2.58 & 0.66 & 7.44 \\
Shades-of-Gray \cite{finlayson2004shades} & 3.22 & 2.12 & 2.44 & 0.43 & 7.77 \\
Cross-dataset CC \cite{koskinen122020cross} & 2.47 & 1.94 & - & - & - \\
Quasi-Unsupervised CC \cite{bianco2019quasi} & 2.69 & 1.76 & 2.00 & 0.49 & 6.45\\
SIIE \cite{afifi2019sensor} & 2.14 & 1.44 & - & 0.44 & 5.06 \\
FFCC \cite{barron2017fast} & 2.69 & 1.89 & 2.08 & 0.46 & 6.31 \\
DMCC \cite{Yue2023EffectiveCC} & 2.23 & 1.63 & 1.78 & 0.49 & 4.95 \\
C5 ($m=7$) \cite{afifi2021cross} & 1.87 & 1.27 & 1.40 & 0.41 & 4.36 \\
C5 ($m=9$) \cite{afifi2021cross} & 1.92 & 1.32 & 1.46 & 0.44 & 4.44 \\
\textbf{CCMNet (Ours)} & \cellcolor[HTML]{\bestcolor}\textbf{1.68} & \cellcolor[HTML]{\bestcolor}\textbf{1.16} & \cellcolor[HTML]{\bestcolor}\textbf{1.26} & \cellcolor[HTML]{\bestcolor}\textbf{0.38} & \cellcolor[HTML]{\bestcolor}\textbf{3.89} \\ 
\bottomrule \\
\end{tabular}}
\vspace{-2mm}

\resizebox{0.95\linewidth}{!}{
\begin{tabular}{lccccc}
\toprule
\textbf{NUS-8 \cite{cheng2014illuminant}} & \textbf{Mean} & \textbf{Med.} & \textbf{Tri.} & \textbf{B.25\%} & \textbf{W.25\%} \\ \midrule \midrule
Gray-world \cite{buchsbaum1980spatial} & 4.59 & 3.46 & 3.81 & 1.16 & 9.85 \\
Shades-of-Gray \cite{finlayson2004shades}& 3.67 & 2.94 & 3.03 & 0.98 & 7.75 \\
Local Surface Reflectance \cite{Gao2014EfficientCC}& 3.45 & 2.51 & 2.70 & 0.98 & 7.32 \\
PCA-based B/W Colors \cite{cheng2014illuminant}& 2.93 & 2.33 & 2.42 & 0.78 & 6.13 \\
Grayness Index \cite{qian2019finding} & 2.91 & 1.97 & 2.13 & 0.56 & 6.67 \\
Cross-dataset CC \cite{koskinen122020cross} & 3.08 & 2.24 & - & - & - \\
Quasi-Unsupervised CC \cite{bianco2019quasi} & 3.00 & 2.25  & - & - & - \\	
FFCC \cite{barron2017fast} & 2.87 & 2.14 & 2.30  & 0.71 & 6.23 \\
C5 ($m=7$) \cite{afifi2021cross} & 2.68 & 2.00 & 2.14 & 0.66 & 5.90 \\
C5 ($m=9$) \cite{afifi2021cross} & 2.54 & 1.90 & 2.02 & 0.61 & 5.61 \\
\textbf{CCMNet (Ours)} & \cellcolor[HTML]{\bestcolor}\textbf{2.32} & \cellcolor[HTML]{\bestcolor}\textbf{1.71} & \cellcolor[HTML]{\bestcolor}\textbf{1.83} & \cellcolor[HTML]{\bestcolor}\textbf{0.53} & \cellcolor[HTML]{\bestcolor}\textbf{5.18} \\ 
\bottomrule \\
\end{tabular}}
\vspace{-2mm}

\resizebox{0.95\linewidth}{!}{
\begin{tabular}{lccccc}
\toprule
\textbf{NUS-8 (CS) \cite{cheng2014illuminant}} & \textbf{Mean} & \textbf{Med.} & \textbf{Tri.} & \textbf{B.25\%} & \textbf{W.25\%} \\ \midrule \midrule
DMCC (CS) \cite{Yue2023EffectiveCC} & 2.80 & 2.12 & 2.25 & 0.74 & 5.88 \\
SIIE (CS) \cite{afifi2019sensor} & 2.05 & 1.50 & - & 0.52 & 4.48  \\
C5 ($m=9$, CS) \cite{afifi2021cross} & 1.77 & 1.37 & 1.46 & 0.48 & 3.75 \\
\textbf{CCMNet (Ours, CS)} & \cellcolor[HTML]{\bestcolor}\textbf{1.71} & \cellcolor[HTML]{\bestcolor}\textbf{1.31} & \cellcolor[HTML]{\bestcolor}\textbf{1.40} & \cellcolor[HTML]{\bestcolor}\textbf{0.48} & \cellcolor[HTML]{\bestcolor}\textbf{3.62} \\
\bottomrule \\
\end{tabular}}
\vspace{-4.5mm}

\caption{Experimental results on three benchmark datasets. CCMNet achieves the best performance across all metrics on various datasets, including additional cross-sensor (CS) validation protocol. For C5 model, $m$ represents the total number of images used, including both the query image and additional images.}
\vspace{-5mm}
\label{tab:main_table}
\end{table}

In contrast, CCMNet achieves superior and more consistent results by leveraging CFE features from two pre-calibrated CCMs. CCMNet is simpler, more robust and does not require retraining for each test camera or the use of additional images. Notably, no data or CCMs from test cameras are used during training, ensuring true zero-shot generalization. Additional visual results are provided in the supplementary materials.

We also report results on the cross-sensor (CS) validation setup \cite{afifi2019sensor}. In this protocol, the network is trained on data from seven cameras in the NUS-8 dataset, excluding one as the test camera. This process is repeated for each of the eight cameras, and the results are averaged.

Following this protocol, we train CCMNet using data from Intel-TAU, Cube+, Gehler-Shi, and seven cameras from NUS-8 (excluding the test camera), aggregating results over eight iterations. As shown at the bottom of \Tref{tab:main_table}, CCMNet outperforms other methods under this evaluation protocol.

Another advantage of CCMNet is its lightweight design. Since the CFE feature is fixed once the camera device is determined, it only needs to be extracted once for a new camera and can be reused thereafter. As a result, CCMNet’s size and computational cost depend solely on the backbone $f$, making it significantly more efficient than the C5 model, which requires 6--8 additional histogram encoders.

As shown in the first table of \Tref{tab:main_table}, the C5 model (with an additional 8 histograms, $m = 9$) requires approximately 2.09 MB of storage, whereas CCMNet, excluding the CFE encoder, occupies only 1.05 MB—almost half the size. This highlights CCMNet’s compactness, making it particularly well-suited for integration into ISP modules, where efficient resource utilization is crucial.

\subsection{Generalization with SIIE Backbone}
\label{sec:generalization}
We further explore using CFE with SIIE \cite{afifi2019sensor}. SIIE learns a 3$\times$3 matrix by processing the raw image $uv$-histogram to map raw colors to a color \textit{working} space. In this experiment, we replace C5 with SIIE as our backbone. Specifically, we input our CFE-concatenated $uv$-histograms, augmented with the imaginary camera transformation, into the SIIE backbone. As shown in \Tref{tab:siie_backbone}, the best performance (MAE) is achieved when both CFE and augmentation are applied, confirming that CCMNet generalizes to different backbones utilizing $uv$-histograms.
\begin{table}
\centering
\resizebox{0.7\linewidth}{!}{
\vspace{-4mm}
\begin{tabular}{lccc}
\toprule
\textbf{Model} & \textbf{Cube+} & \textbf{Gehler-Shi} & \textbf{NUS-8} \\ \midrule \midrule
 
SIIE \cite{afifi2019sensor} & 3.39 & 3.67 & 3.52 \\
w/ CFE & 2.60 & 3.62 & 3.36 \\
w/ aug. & 2.43 & 3.12 & 3.00 \\
w/ CFE \& aug. & \cellcolor[HTML]{\bestcolor}\textbf{1.91} & \cellcolor[HTML]{\bestcolor}\textbf{2.99} & \cellcolor[HTML]{\bestcolor}\textbf{2.94} \\
\bottomrule
\end{tabular}}
\vspace{-2mm}
\caption{Generalization with the SIIE \cite{afifi2019sensor} backbone. Reported results show the mean angular error.}
\label{tab:siie_backbone}
\end{table}

\subsection{Ablation Studies}
\label{sec:ablation}
\begin{table}
\centering
\resizebox{0.89\linewidth}{!}{
\begin{tabular}{lcccc}
\toprule
\textbf{Model} & \textbf{Aug. method} & \textbf{Cube+} & \textbf{Gehler-Shi} & \textbf{NUS-8} \\ \midrule \midrule
\multirow{3}{*}{Backbone $f$} 
& w/o aug. & 2.22 & 2.79 & 2.88 \\
& $\alpha=1$ & 1.94 & 2.87 & 2.50 \\
& $0\leq\alpha\leq1$ & 1.78 & 2.53 & 2.54 \\ \cmidrule(lr){1-5}
\multirow{3}{*}{\makecell{CCMNet \\ ($f$ + CFE)}}
& w/o aug. & 2.23 & 2.74 & 2.70 \\
& $\alpha=1$ & 1.86 & 2.34 & 2.45 \\
& $0\leq\alpha\leq1$ & \cellcolor[HTML]{\bestcolor}\textbf{1.68} & \cellcolor[HTML]{\bestcolor}\textbf{2.23} & \cellcolor[HTML]{\bestcolor}\textbf{2.32} \\
\bottomrule
\end{tabular}}
\vspace{-2mm}
\caption{Ablation studies on the impact of the CFE encoder and different augmentation strategies. The reported results are the mean angular error.}
\label{tab:ablation}
\vspace{-5mm}
\end{table}
\Tref{tab:ablation} presents the performance of the backbone $f$ and CCMNet trained under three setups: without augmentation (w/o aug), with augmentation at $\alpha=1$, and with augmentation for $0 \leq \alpha \leq 1$. Architecturally, the backbone $f$ mirrors the $m=1$ structure from C5 \cite{afifi2021cross}, excluding additional images and encoders. Setting $\alpha=1$ in the augmentation process replicates the camera-mapping strategy used in C5.

The training data is halved for the w/o aug. setup, and the number of iterations is doubled to ensure the same number of model updates as in the other experiments. The results indicate that the CFE encoder in CCMNet and the imaginary camera augmentation play crucial roles in the cross-camera color constancy task.

\section{Conclusion and Discussion}
In this paper, we propose CCMNet, a lightweight and efficient method for cross-camera color constancy that leverages pre-calibrated CCMs available in camera ISPs. The model utilizes these CCMs, which map a camera’s raw color space to the device-independent CIE XYZ color space or vice versa, to encode the camera-specific illumination locus into a guidance embedding. This feature, termed CFE, directs a hypernetwork to quickly adapt to unseen cameras during testing, enabling the generation of appropriate filters and biases while achieving superior performance compared to previous methods.

By taking advantage of the linearity of CCM operations, the proposed imaginary camera augmentation technique allows the model to learn a broader range of virtual camera response functions during training, significantly improving CCMNet’s generalization capability.

While most cameras include calibrated raw-to-XYZ CCMs in their ISPs and DNG files, some smartphones may not provide accurate CCMs in their DNGs. Instead, these devices often include a single fixed matrix to convert raw images to linear sRGB. This limitation could hinder our method’s ability to process DNG files from such devices or necessitate an additional conversion step to adapt to the raw-to-linear sRGB matrix.
\clearpage
\setcounter{page}{9}
\maketitlesupplementary
\appendix

\section{CCMs \& CCTs Extraction}\label{sec:a}

In this section, we describe the methodology used to extract the color correction matrices, low and high correlated color temperatures ($\textrm{CCT}_{low}$, $\textrm{CCT}_{high}$) information utilized in our approach. Since CCMs and their correlated CCTs are camera-dependent, they can be extracted once and remain consistent across all images captured by the same camera.

To extract the CCMs and CCTs of a specific camera, we followed these steps. First, to ensure consistency in data processing, we converted all raw images to the DNG format using Adobe DNG Converter, instead of relying on camera-specific raw file extensions. Second, we extracted metadata from the DNG files using ExifTool, specifically retrieving \textit{ColorMatrix1}, \textit{ColorMatrix2}, \textit{ForwardMatrix1}, and \textit{ForwardMatrix2}. These matrices were then used for our imaginary camera augmentation and for testing on previously unseen cameras during training.
For convenience, we will refer to \textit{ColorMatrix} and \textit{ForwardMatrix} as CM and FM, respectively, throughout this supplementary material. 

\begin{figure}
\centering
\includegraphics[width=\linewidth]{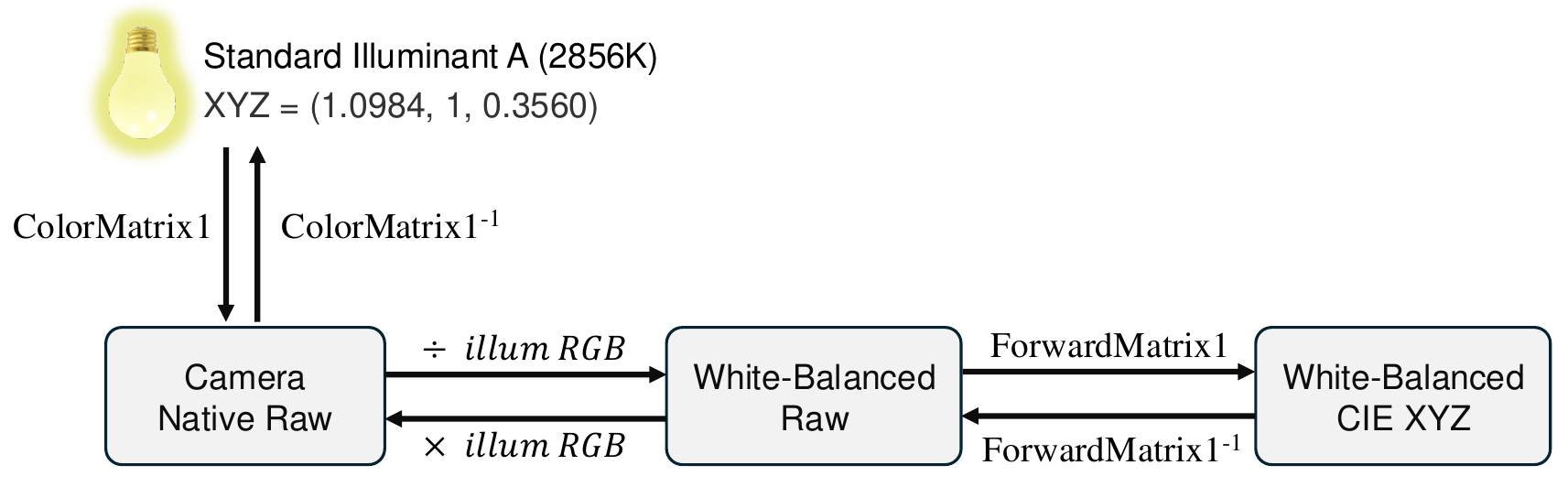}
\caption{A schematic diagram illustrating the use of \textit{ColorMatrix} and \textit{ForwardMatrix}. The \textit{ForwardMatrix} (FM) transforms white-balanced raw data into the CIE XYZ color space, while the \textit{ColorMatrix} (CM) converts CIE XYZ values of a standard light source into the camera's native raw color space. FM1 and CM1 are calibrated for standard illuminant A (2856K), and FM2 and CM2 are calibrated for the D65 illuminant (6504K).}
\vspace{-1em}
\label{fig:ccm}
\end{figure}

\Fref{fig:ccm} illustrates the relationships between color spaces and the transformation matrices involved. As shown, the FM transforms white-balanced camera raw colors to the CIE XYZ color space, while the CM converts from CIE XYZ to the camera’s native raw color space under a specific illuminant. The suffixes `1' and `2' in the matrix names indicate calibration for illuminants 1 and 2, corresponding to standard illuminant A and D65, respectively. Accordingly, we define $\textrm{CCT}_{low}$ and $\textrm{CCT}_{high}$ as the color temperatures of illuminant A (2856K) and D65 (6504K) and use these values for CCM interpolation, as described in \Eref{eq10:ccm_interp} in the main paper.

As defined in \Eref{eq9:xyz2raw} in the main paper, the $\textrm{CCM}_{low}$ and $\textrm{CCM}_{high}$ matrices used throughout this work correspond to CM1 and CM2, respectively. Additionally, CM1, CM2, FM1, FM2 are used in the imaginary camera augmentation process described in \Sref{sec:c2c}.

\section{Details of the CFE Encoding Process} \label{sec:b}

\begin{figure*}
\centering
\includegraphics[width=\linewidth]{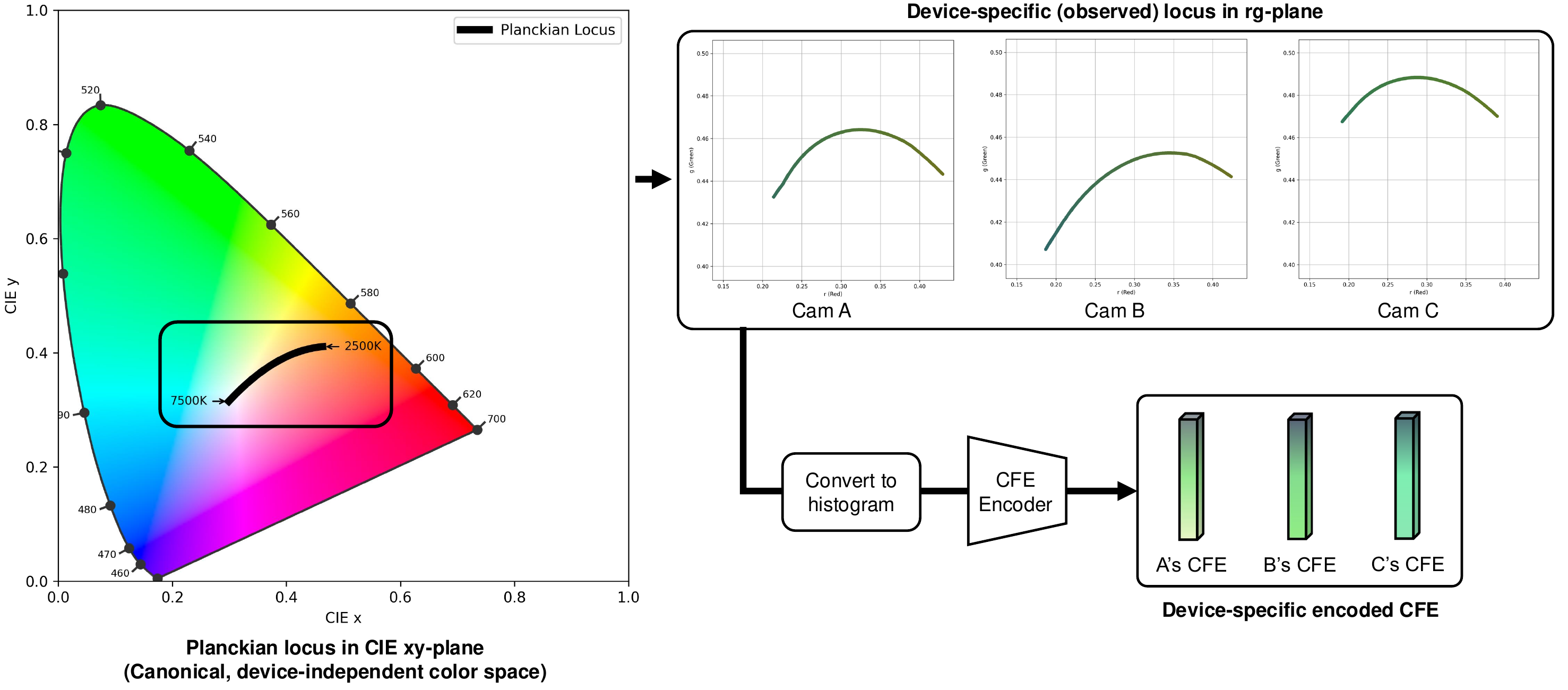}
\caption{Detailed visualization of CFE encoding process. As mentioned in the main paper, the camera’s \textit{fingerprint} is derived by converting the reference CIE XYZ colors (locus) along the correlated color temperature (CCT) range of 2500K–7500K into the corresponding RGB locus as \textit{observed} by each device, followed by an encoding process. Due to this characteristic, the CFE feature inherently reflects the color characteristics induced by each camera’s spectral sensitivity.}
\vspace{-1em}
\label{fig:locus}
\end{figure*}

In this section, we provide additional details on the CFE (Camera Fingerprint Embedding) encoding process described in \Sref{sec:cfe}.

\paragraph{Further explanations.}
As shown in \Fref{fig:locus}, the essence of what CFE fundamentally encodes is the color trajectory on the CIE xy-plane within the correlated color temperature (CCT) range of 2500K–7500K. These colors correspond to the light emitted by a black body at a given CCT and are intrinsic, invariant values. However, due to differences in the spectral sensitivity of imaging sensors, each device \textit{observes} these reference colors as distinct loci. These trajectories inherently represent the unique color characteristics of each device.

We leverage the fact that this \textit{observation} process is precomputed for two illuminants during the ISP manufacturing stage and recorded as matrices (CCMs). By interpolating the two matrices, $\textrm{CCM}_{low}$ and $\textrm{CCM}_{high}$, and then applying to the Planckian XYZ locus, we replace the observation process for each device. The resulting device-specific locus is then converted into a histogram, which is subsequently encoded into a CFE feature that captures the \textit{fingerprint} of each camera using a CNN-based CFE encoder.

Due to this design approach of the CFE feature, the CCMNet leverages CFE as guidance, enabling it to infer and adapt to the color space of a previously unseen camera. This allows the model to learn a generalized approach to illuminant color estimation without requiring explicit training on every individual camera.

\paragraph{Technical details.}
For the XYZ locus corresponding to color temperatures from 2500K to 7500K, we used the \texttt{colour.temperature.CCT\_to\_xy} function from the \texttt{colour} Python library. A total of 51 chromaticity coordinates were sampled at 100K intervals, ranging from 2500K to 7500K.

As mentioned in the main paper, the sampled XYZ locus was transformed into the camera's native raw RGB space by interpolating between CM1 and CM2. This was further converted into a histogram with 64 bins, within the $uv$ range of [-0.5, 1.5]. The resulting 64$\times$64$\times$1 histogram was processed by the CFE encoder, which outputs an 8-dimensional embedding vector. The CFE encoder consists of four \texttt{DoubleConvBlocks} followed by a projection head. Each \texttt{DoubleConvBlock} processes the input by applying two convolutional layers, each with a kernel size of $3\times3$, a stride of 1, and a padding of 1, followed by a LeakyReLU activation. This is then followed by a $2\times2$ max-pooling layer and batch normalization. The projection head flattens the feature map and maps it to an 8-dimensional embedding vector using an MLP with two hidden layers.
\section{Camera-to-Camera Mapping} \label{sec:c2c}
\begin{figure*}
\centering
\includegraphics[width=\linewidth]{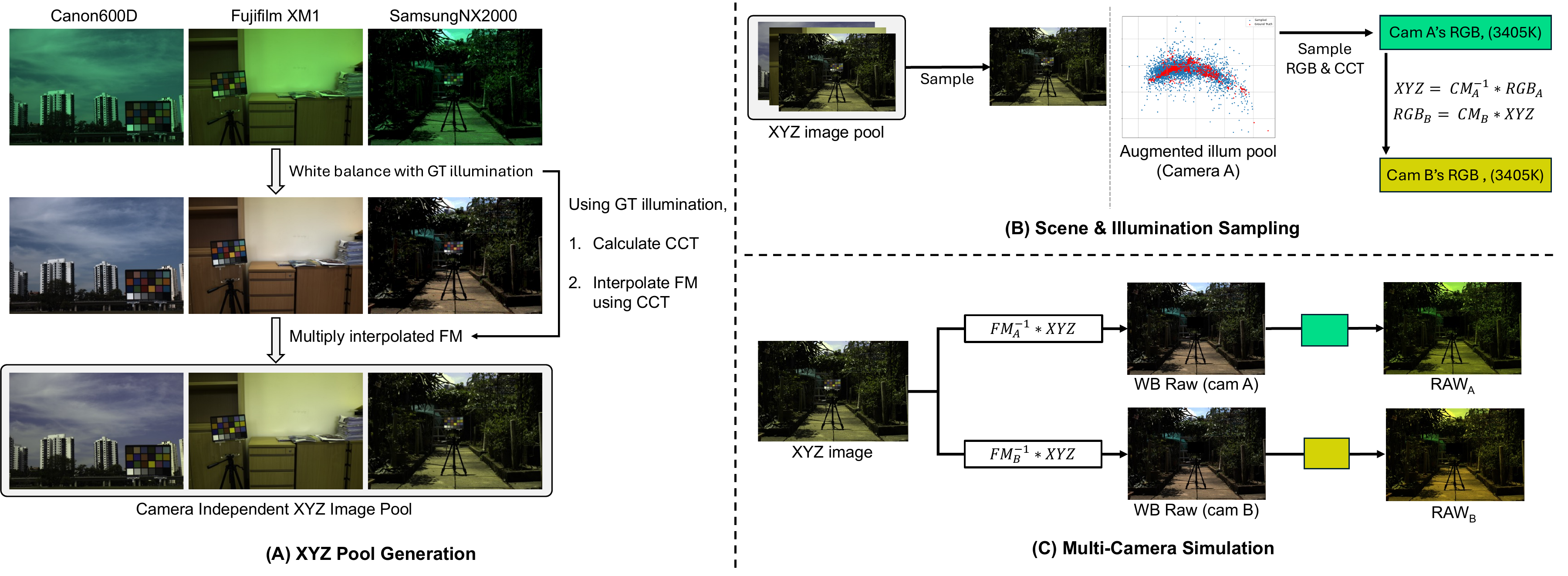}
\caption{Overall process of camera-to-camera mapping. In (A), subsets of images taken by different cameras from multiple datasets are white-balanced using the corresponding ground-truth illuminants, and the \textit{ForwardMatrix} is used to convert them to the CIE XYZ space, creating the XYZ image pool. In (B), a reference image is sampled from the pool, and an illumination color is sampled from the augmented illumination pool of the source camera (Camera A) that originally captured the image. The sampled illumination is then mapped to the native RGB space of a randomly selected target camera (Camera B) using the \textit{ColorMatrix}. Finally, in (C), the XYZ image is transformed into the white-balanced color space of Cameras A and B using the inverse of their respective \textit{ForwardMatrices}, and illumination casting is applied by multiplying the images with the illumination RGB values of each camera space.}
\vspace{-1em}
\label{fig:c2c}
\end{figure*}

In Sec.~\ref{sec:augmentation} of the main paper, we introduced our imaginary camera augmentation, which assumes two versions of the same image in the camera's native raw RGB space. To satisfy this condition, we perform a camera-to-camera mapping inspired by \cite{afifi2021cross}. In this section, we provide a detailed explanation of the camera-to-camera mapping process used in our work. Specifically, in \Sref{sec:c1}, we explain the process of computing the correlated color temperature (CCT) of a light source in the target camera's native raw RGB space. Then, in \Sref{sec:c2}, we describe how to generate a pool of white-balanced, camera-independent XYZ images using the RGB values of the light source and the corresponding CCT. In \Sref{sec:c3}, we describe the process of generating a device-specific illumination pool for random sampling. Finally, \Sref{sec:c4} explains our camera-to-camera mapping, which presents a reference image in two different camera-native raw RGB spaces. The reference image is sampled from the XYZ image pool, while the illumination is sampled from the augmented ground-truth (GT) illumination pool of each camera. The overall process is visualized in \Fref{fig:c2c}.

While our camera-to-camera mapping is inspired by the C5 augmentation approach \cite{afifi2021cross}, it differs in the following ways. First, we remove C5’s restriction that limits sampling from the illumination pool to similar scenes with matching capturing settings (e.g., ISO, exposure time) and illumination CCT. Specifically, in C5, both the sampled scene image from the CIE XYZ space and the sampled illuminant from the target camera were required to have similar capturing settings and CCT. In contrast, our approach removes this constraint, eliminating the need to rely on capturing settings and allowing for greater diversity in augmentation. Additionally, instead of sampling from a fitted cubic polynomial based on the target camera's illuminant samples, we use a fitted cubic polynomial based on the illuminant values from the source camera’s dataset (i.e., the camera from which the reference XYZ image was taken). The sampled illuminant is then transferred to the CIE XYZ space using the inverse of the source camera's CM, followed by a transformation of these CIE XYZ illuminant values into the native raw RGB space of the target camera.

\subsection{Illumination RGB to CCT Conversion} \label{sec:c1}
The illuminant estimation datasets used in the main paper provide GT illumination RGB labels for each scene in the camera’s native raw RGB space. According to the Adobe DNG specification, given CM1 and CM2 (extracted for each camera as described in \Sref{sec:a}), along with the GT illumination RGB, the CCT and CIE XYZ values of the light source can be computed using Algorithm \ref{alg:camneutral_to_xyz}.

\begin{algorithm}
\caption{Conversion of Illuminant Raw RGB to CCT and XYZ Coordinates}
\label{alg:camneutral_to_xyz}
\begin{algorithmic}[1]
\Function{camntrl\_to\_xyz}{illum, cm1, cm2}
    \State xy = [0.3127, 0.3290]
    \While{True}
        \State cct = colour.temperature.xy\_to\_CCT(xy)
        \State color\_matrix = interpolate\_ccm(cct, cm1, cm2)
        \State color\_matrix\_inv = np.linalg.inv(color\_matrix)
        \State xyz = np.dot(color\_matrix\_inv, illum)
        \State X, Y, Z = xyz
        \State xy\_new = [X / (X + Y + Z), Y / (X + Y + Z)]
        \If{np.allclose(xy, xy\_new, atol=1e-6)}
            \State \Return xyz, cct
        \EndIf
        \State xy = xy\_new
    \EndWhile
\EndFunction
\end{algorithmic}
\end{algorithm}

The algorithm iteratively estimates the CCT and converts illuminant RGB values to the CIE XYZ space. Using metadata such as CM1 and CM2, it interpolates the appropriate color correction matrix for the estimated CCT and applies it to transform the input illumination into the CIE XYZ space. The resulting XYZ coordinates and CCT values are then used either to generate the camera-independent XYZ image pool in \Sref{sec:c2} or to transform the illumination into the target camera RGB space in \Sref{sec:c4}.

\subsection{Unified XYZ Image Pool Generation} \label{sec:c2}
In this section, we describe the process of creating an XYZ image pool for camera-to-camera mapping by converting images captured by various cameras into the device-independent XYZ color space. The process involves two main steps: (1) white balancing with GT labels, and (2) transforming to the CIE XYZ color space using the \textit{ForwardMatrix} (FM). Refer to \Fref{fig:ccm} and \Fref{fig:c2c}-(A).

As explained in the main paper, we use multiple datasets captured by various cameras, each including GT illumination labels that enable accurate white balancing of images in the camera's native raw RGB space. As described in \Sref{sec:a}, we extract FM1 and FM2 for each camera. Using the CCT of the GT illumination, we interpolate between FM1 and FM2 to transform the white-balanced images into the XYZ color space. The CCT is computed from the GT illumination RGB using the method detailed in \Sref{sec:c1}.

This process mitigates the dependency on camera specifications, and in theory, the images are independent of camera models and illumination conditions. By aggregating these images, we construct a unified XYZ image pool that serves as the foundation for camera-to-camera mapping.

\subsection{Camera-specific Illumination Pool Generation} \label{sec:c3}
Next, we generate an illumination pool for each camera. While it is possible to use only the GT illuminations, we adopt the augmentation method proposed in \cite{afifi2021cross} to enhance generality and diversity. This method involves fitting a cubic polynomial to the GT illuminations for each camera and then introducing random shifts to augment the illuminations. For further details, please refer to the supplementary material of \cite{afifi2021cross}. On the right side of \Fref{fig:c2c}-(B), we show a plot of the illumination pool for a specific camera (Camera A). In this plot, the red points represent the GT illumination labels extracted from the dataset, while the blue points correspond to the augmented illuminations.

\subsection{Camera-to-Camera Image Synthesis}\label{sec:c4}
In this section, we describe a camera-to-camera mapping method that simulates the same scene as if it were captured by two different cameras, using the image pool from \Sref{sec:c2} and the illumination pool from \Sref{sec:c3}. See \Fref{fig:c2c}-(B) and (C).

\paragraph{Scene and Illumination Sampling \& Mapping.} First, a scene is randomly selected from the XYZ image pool. Next, an illumination is randomly sampled from the illumination pool of the source camera that captured the selected scene. This sampled illumination is then transformed into the native raw color space of a randomly selected target camera from the set of cameras used (see \Fref{fig:c2c}-(B)). As illustrated in \Fref{fig:ccm}, the XYZ values of the sampled illumination are computed by applying the inverse of the source camera's \textit{ColorMatrix} (CM). These XYZ values are then multiplied by the target camera’s CM to obtain the native raw color of the illumination in the target camera’s color space. The interpolation of each camera’s CM is based on the CCT of the illumination, which is calculated using the steps described in \Sref{sec:c1}.

\paragraph{Synthesizing Paired Scene from Two Cameras.}
Finally, as illustrated in \Fref{fig:c2c}-(C), we generate two raw images of the sampled scene, as if it were captured by the selected two cameras under the same sampled illumination. As shown in \Fref{fig:ccm}, the white-balanced XYZ image is transferred to the cameras' native raw space in two steps. First, using the same CCT employed during CM interpolation in illumination mapping, the FMs of cameras A and B are interpolated, and their inverses are applied to the XYZ image. This step produces two white-balanced raw images, one for each camera. Next, the camera-native illumination RGB values--sampled from camera A and mapped to camera B as described in previous paragraph--are multiplied with these raw images. The resulting image pair simulates the same scene and lighting conditions as captured by two different cameras, all derived from a single XYZ image.
\section{Imaginary Camera Augmentation Visualizations}
\begin{figure*}
\centering
\includegraphics[width=\linewidth]{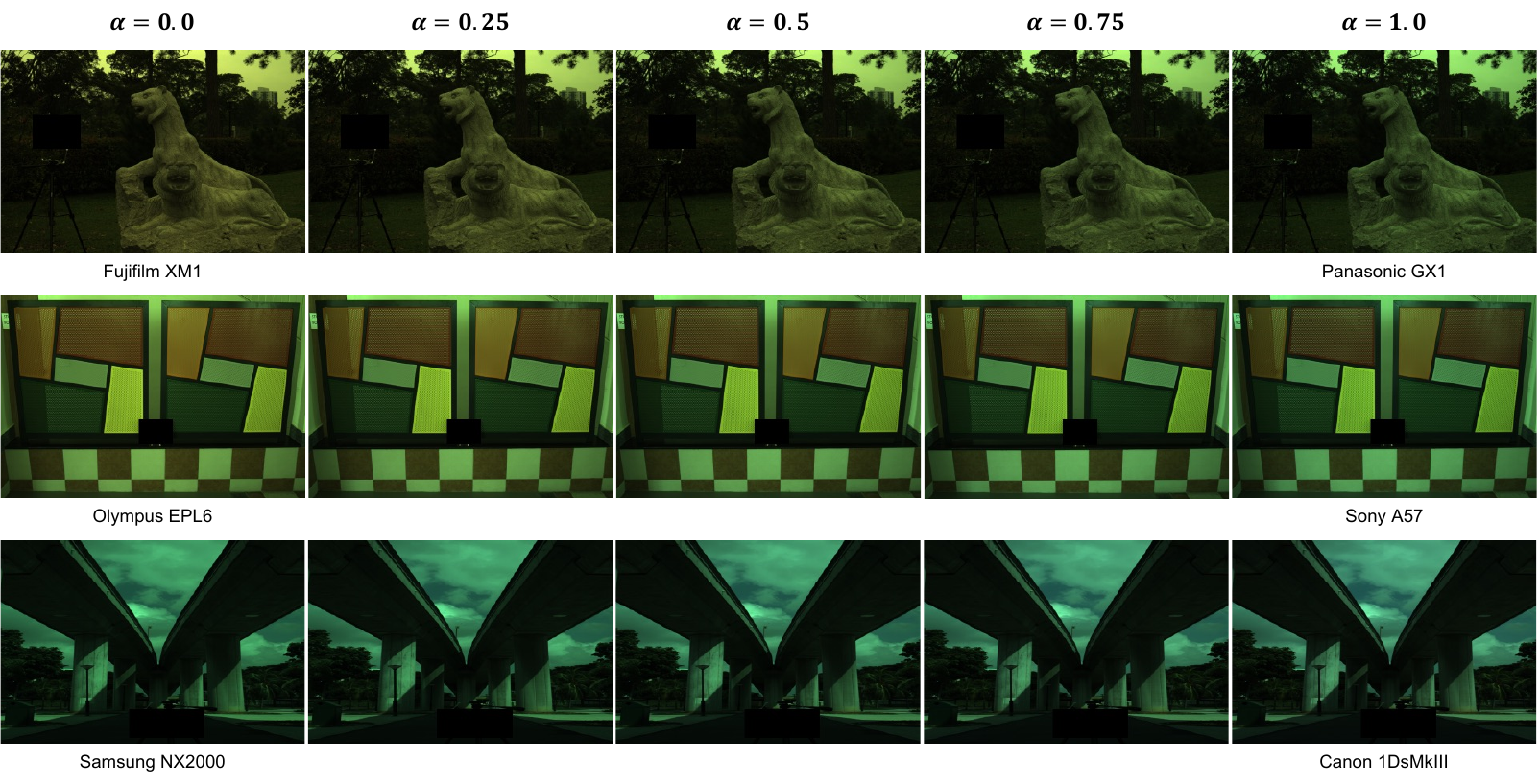}
\caption{Results of our imaginary camera augmentation. In each row, the leftmost and rightmost images represent the source and target camera images generated using the method described in \Sref{sec:c2c}, while the three middle images represent those produced by the \textit{imaginary} camera, generated by interpolating between the two devices at ratios of 0.25, 0.5, and 0.75, respectively. As explained in \Sref{sec:augmentation} of the main paper, the CCMs of the imaginary cameras are interpolated using the same alpha values applied during image interpolation, and the resulting CFE embeddings are generated for training. Brightness is adjusted for visibility.}
\label{fig:imaginary}
\end{figure*}

Here, we provide additional visualizations of the imaginary camera augmentation. As shown in \Fref{fig:imaginary}, Imaginary Camera Augmentation simulates images captured by virtual cameras that interpolate the properties of two real-world devices. This data augmentation technique also interpolates the CCMs at the same ratios to generate the CCMs for these virtual cameras.

\section{Experimental Setup} \label{sec:e}

As mentioned in the main paper, the backbone $f$ uses the standard U-Net-like architecture from C5 \cite{afifi2021cross}. However, unlike C5, we do not use additional images from the test camera, so no extra encoders are employed. Instead, we use a single Encoder-Decoder U-Net architecture. The encoder and decoder are connected via skip connections, with each consisting of four \texttt{DoubleConv} layers. In the encoder, each \texttt{DoubleConv} layer is followed by max pooling, while in the decoder, feature upsampling and skip connections are applied before each \texttt{DoubleConv} layer.

The batch size was set to 16, and training was conducted over 50 epochs with an initial learning rate of $5 \times 10^{-4}$. A learning rate decay of 0.5 was applied at epoch 25. The Adam optimizer \cite{kingma2014adam} was used for training.

For data augmentation, camera-to-camera mapping and imaginary camera augmentation are applied exclusively using the camera subsets from the training datasets, excluding the test dataset. For instance, when evaluating the Cube+ dataset, the augmented dataset used for model training is generated from images and CCMs from the Gehler-Shi \cite{shi2000re}, NUS-8 \cite{cheng2014illuminant}, and Intel-TAU \cite{laakom2021intel} datasets.

\section{Additional Results} \label{sec:additional}

\begin{figure*}
\centering
\includegraphics[width=\linewidth]{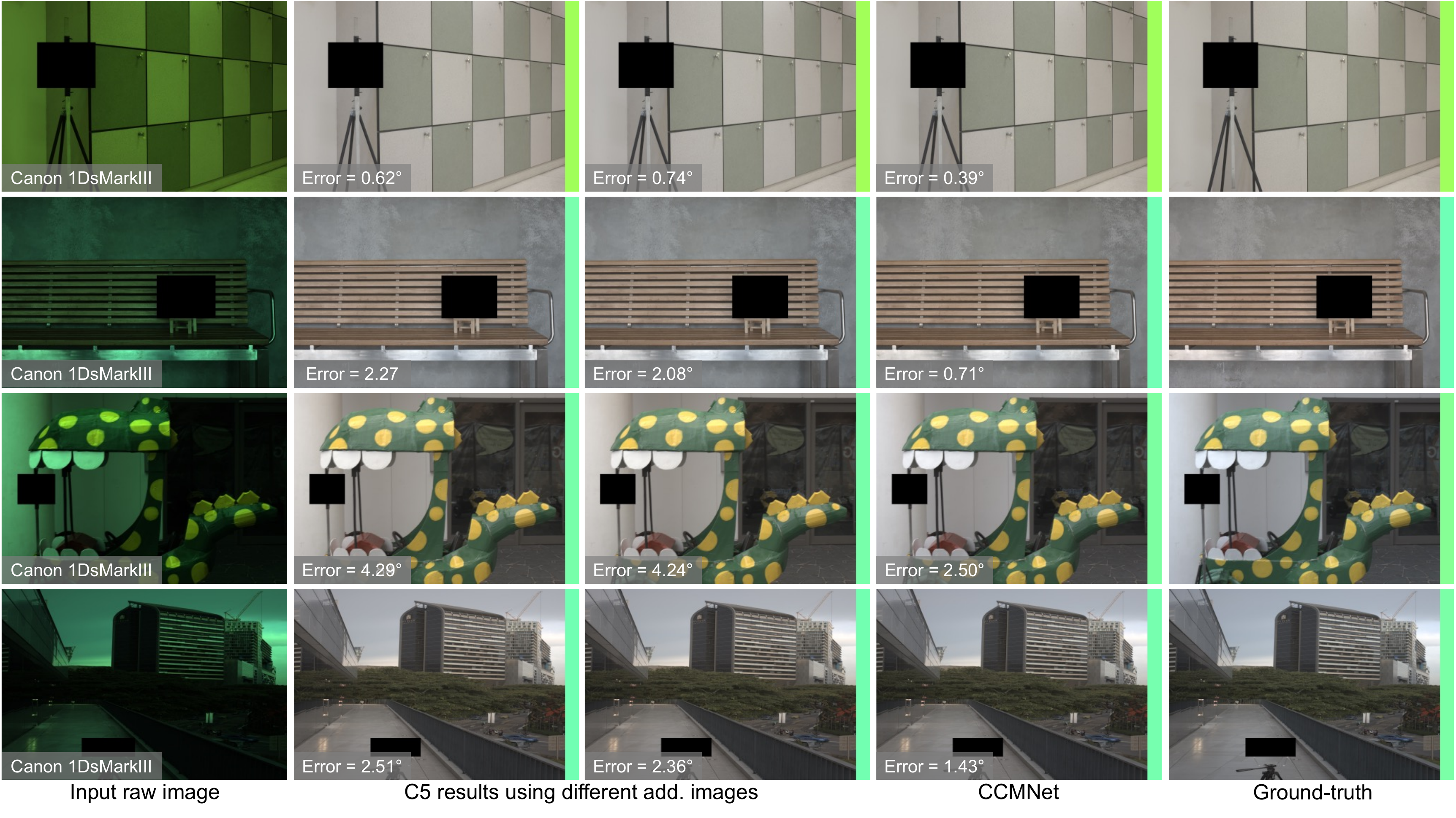}
\caption{Additional results for Canon EOS 1Ds Mark III. CCMNet demonstrates superior performance on various scenes captured by unseen camera. 
Notably, CCMNet has never been exposed to any images or the CCM of the Canon 1Ds Mark III during training.}
\label{fig:add1}
\end{figure*}

\begin{figure*}
\centering
\includegraphics[width=\linewidth]{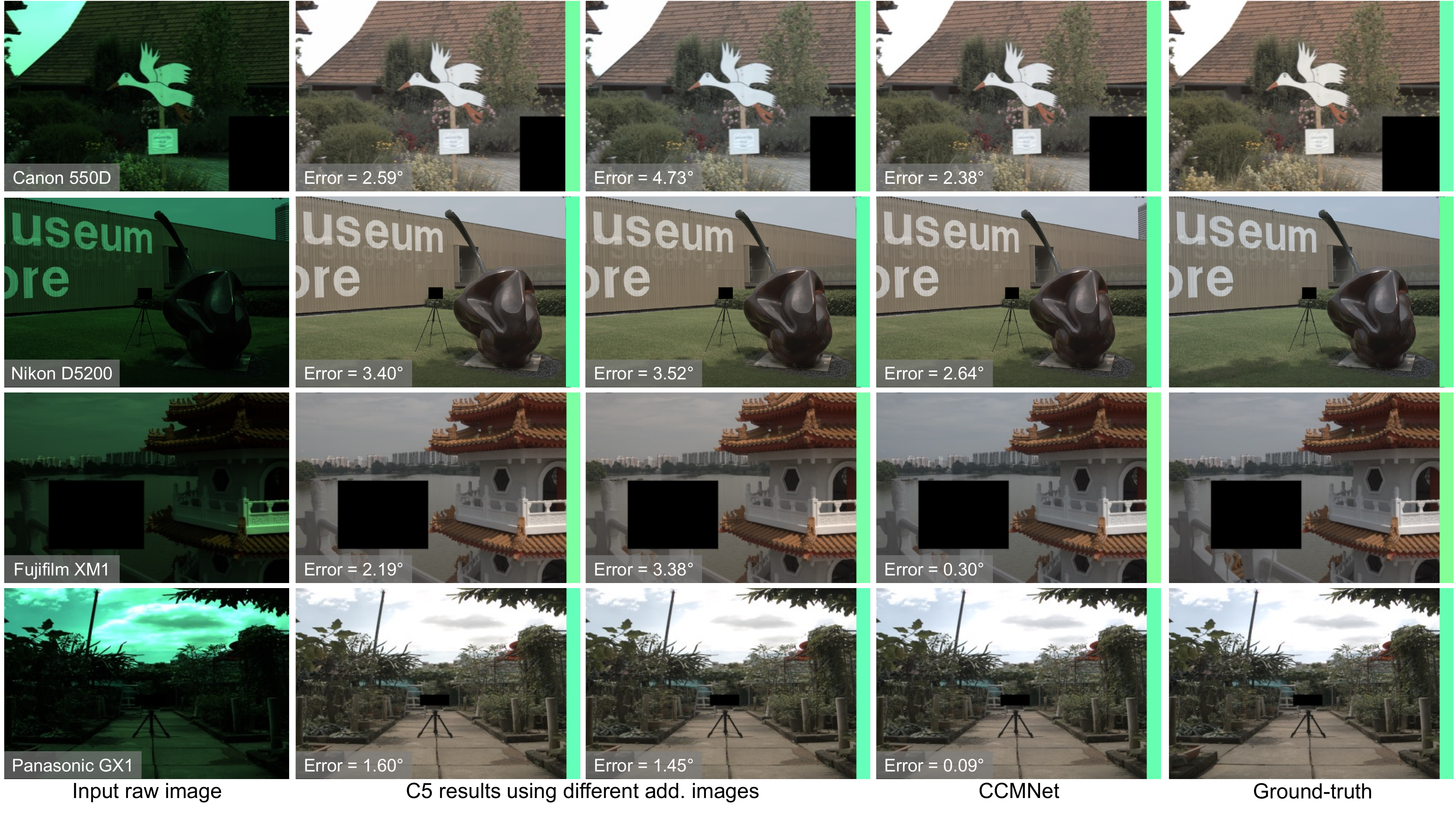}
\caption{Additional results for various cameras show that CCMNet exhibits robust performance across a range of unseen cameras. Importantly, it has not been exposed to any images or CCMs from the cameras shown in the figure during training.}
\label{fig:add2}
\end{figure*}

We present additional visualization results in \Fref{fig:add1} and \Fref{fig:add2}. As shown in \Fref{fig:add1}, CCMNet achieves satisfactory accuracy across various scenes captured by a camera it has never encountered during training. In \Fref{fig:add2}, we demonstrate that CCMNet maintains robust accuracy across a set of unseen cameras.

\newpage
{
    \small
    \bibliographystyle{ieeenat_fullname}
    \bibliography{main}

@String(CVPR= {IEEE Conf. Comput. Vis. Pattern Recog.})

@String(ICCV= {Int. Conf. Comput. Vis.})

@String(ECCV= {Eur. Conf. Comput. Vis.})

@String(BMVC= {Brit. Mach. Vis. Conf.})

@String(TOG= {ACM Trans. Graph.})

@String(ICIP = {IEEE Int. Conf. Image Process.})

@String(AAAI = {AAAI})

@String(CVPRW= {IEEE Conf. Comput. Vis. Pattern Recog. Worksh.})

@String(CVPR  = {CVPR})

@String(ICCV  = {ICCV})

@String(ECCV  = {ECCV})

@String(BMVC  =	{BMVC})

@String(TOG   = {ACM TOG})

@String(ICIP  = {ICIP})

@String(CVPRW= {CVPRW})

@inproceedings{gehler2008bayesian,
  title={Bayesian color constancy revisited},
  author={Gehler, Peter Vincent and Rother, Carsten and Blake, Andrew and Minka, Tom and Sharp, Toby},
  booktitle={CVPR},
  year={2008},
}

@article{shi2000re,
  title={Re-processed version of the gehler color constancy dataset of 568 images},
  author={Shi, Lilong},
  journal={http://www. cs. sfu. ca/\~{} color/data/},
  year={2000}
}

@article{cheng2014illuminant,
  title={Illuminant estimation for color constancy: {W}hy spatial-domain methods work and the role of the color distribution},
  author={Cheng, Dongliang and Prasad, Dilip K and Brown, Michael S},
  journal={JOSA A},
  volume={31},
  number={5},
  pages={1049--1058},
  year={2014},
}

@article{banic2017unsupervised,
  title={Unsupervised learning for color constancy},
  author={Bani{\'c}, Nikola and Ko{\v{s}}{\v{c}}evi{\'c}, Karlo and Lon{\v{c}}ari{\'c}, Sven},
  journal={arXiv preprint arXiv:1712.00436},
  year={2017}
}

@article{laakom2021intel,
  title={Intel-tau: A color constancy dataset},
  author={Laakom, Firas and Raitoharju, Jenni and Nikkanen, Jarno and Iosifidis, Alexandros and Gabbouj, Moncef},
  journal={IEEE access},
  volume={9},
  pages={39560--39567},
  year={2021},
  publisher={IEEE}
}

@article{buchsbaum1980spatial,
  title={A spatial processor model for object colour perception},
  author={Buchsbaum, Gershon},
  journal={Journal of the Franklin institute},
  volume={310},
  number={1},
  pages={1--26},
  year={1980},
}

@article{land1977retinex,
  title={The retinex theory of color vision},
  author={Land, Edwin H},
  journal={Scientific american},
  volume={237},
  number={6},
  pages={108--129},
  year={1977},
}

@article{van2007edge,
  title={Edge-based color constancy},
  author={Van De Weijer, Joost and Gevers, Theo and Gijsenij, Arjan},
  journal={IEEE Transactions on image processing},
  volume={16},
  number={9},
  pages={2207--2214},
  year={2007},
}

@inproceedings{finlayson2004shades,
  title={Shades of gray and colour constancy},
  author={Finlayson, Graham D and Trezzi, Elisabetta},
  booktitle={Color and Imaging Conference},
  year={2004},
}

@article{kingma2014adam,
  title={Adam: A method for stochastic optimization},
  author={Kingma, Diederik P},
  journal={arXiv preprint arXiv:1412.6980},
  year={2014}
}

@article{gijsenij2010generalized,
  title={Generalized gamut mapping using image derivative structures for color constancy},
  author={Gijsenij, Arjan and Gevers, Theo and Van De Weijer, Joost},
  journal={International Journal of Computer Vision},
  volume={86},
  number={2-3},
  pages={127--139},
  year={2010},
}

@article{gijsenij2011improving,
  title={Improving color constancy by photometric edge weighting},
  author={Gijsenij, Arjan and Gevers, Theo and Van De Weijer, Joost},
  journal={IEEE Transactions on Pattern Analysis and Machine Intelligence},
  volume={34},
  number={5},
  pages={918--929},
  year={2011},
}

@article{qian2018revisiting,
  title={Revisiting gray pixel for statistical illumination estimation},
  author={Qian, Yanlin and Pertuz, Said and Nikkanen, Jarno and K{\"a}m{\"a}r{\"a}inen, Joni-Kristian and Matas, Jiri},
  journal={arXiv preprint arXiv:1803.08326},
  year={2018}
}

@inproceedings{qian2019finding,
  title={On finding gray pixels},
  author={Qian, Yanlin and Kamarainen, Joni-Kristian and Nikkanen, Jarno and Matas, Jiri},
  booktitle={CVPR},
  year={2019}
}

@article{ulucan2024multi,
  title={Multi-scale color constancy based on salient varying local spatial statistics},
  author={Ulucan, Oguzhan and Ulucan, Diclehan and Ebner, Marc},
  journal={The Visual Computer},
  volume={40},
  number={9},
  pages={5979--5995},
  year={2024},
}

@article{oh2017approaching,
  title={Approaching the computational color constancy as a classification problem through deep learning},
  author={Oh, Seoung Wug and Kim, Seon Joo},
  journal={Pattern Recognition},
  volume={61},
  pages={405--416},
  year={2017},
}

@inproceedings{barron2017fast,
  title={Fast {F}ourier color constancy},
  author={Barron, Jonathan T and Tsai, Yun-Ta},
  booktitle={CVPR},
  year={2017}
}

@inproceedings{barron2015convolutional,
  title={Convolutional color constancy},
  author={Barron, Jonathan T},
  booktitle={ICCV},
  year={2015}
}

@inproceedings{bianco2015color,
  title={Color constancy using {CNN}s},
  author={Bianco, Simone and Cusano, Claudio and Schettini, Raimondo},
  booktitle={CVPRW},
  year={2015}
}

@inproceedings{shi2016deep,
  title={Deep specialized network for illuminant estimation},
  author={Shi, Wu and Loy, Chen Change and Tang, Xiaoou},
  booktitle={ECCV},
  year={2016},
}

@article{li2023ranking,
  title={Ranking-based color constancy with limited training samples},
  author={Li, Bing and Qin, Haina and Xiong, Weihua and Li, Yangxi and Feng, Songhe and Hu, Weiming and Maybank, Stephen},
  journal={IEEE Transactions on Pattern Analysis and Machine Intelligence},
  volume={45},
  number={10},
  pages={12304--12320},
  year={2023},
}

@inproceedings{koskinen122020cross,
  title={Cross-dataset color constancy revisited using sensor-to-sensor transfer},
  author={Koskinen12, Samu and Yang, Dan and K{\"a}m{\"a}r{\"a}inen, Joni-Kristian},
  year={2020},
  organization={BMVC}
}

@article{Yue2023EffectiveCC,
  title={Effective cross-sensor color constancy using a dual-mapping strategy.},
  author={Shuwei Yue and Minchen Wei},
  journal={Journal of the Optical Society of America. A, Optics, image science, and vision},
  year={2023},
  volume={41 2},
  pages={
          329-337
        },
  url={https://api.semanticscholar.org/CorpusID:266317702}
}

@inproceedings{afifi2021cross,
  title={Cross-camera convolutional color constancy},
  author={Afifi, Mahmoud and Barron, Jonathan T and LeGendre, Chloe and Tsai, Yun-Ta and Bleibel, Francois},
  booktitle={ICCV},
  year={2021}
}

@misc{hubel2007white,
  title={White point estimation using color by convolution},
  author={Hubel, Paul M and Finlayson, Graham D and Hordley, Steven D},
  year={2007},
  note={US Patent 7,200,264}
}

@article{finlayson2001color,
  title={Color constancy at a pixel},
  author={Finlayson, Graham D and Hordley, Steven D},
  journal={JOSA A},
  volume={18},
  number={2},
  pages={253--264},
  year={2001},
}

@inproceedings{fourure2016mixed,
  title={Mixed pooling neural networks for color constancy},
  author={Fourure, Damien and Emonet, R{\'e}mi and Fromont, Elisa and Muselet, Damien and Tr{\'e}meau, Alain and Wolf, Christian},
  booktitle={ICIP},
  year={2016},
}

@inproceedings{lou2015color,
  title={Color Constancy by Deep Learning.},
  author={Lou, Zhongyu and Gevers, Theo and Hu, Ninghang and Lucassen, Marcel P and others},
  booktitle={BMVC},
  year={2015}
}

@inproceedings{abdelhamed2021leveraging,
  title={Leveraging the availability of two cameras for illuminant estimation},
  author={Abdelhamed, Abdelrahman and Punnappurath, Abhijith and Brown, Michael S},
  booktitle={CVPR},
  year={2021}
}

@article{Afifi2020CIEXN,
  title={{CIE XYZ} {N}et: Unprocessing Images for Low-Level Computer Vision Tasks},
  author={Mahmoud Afifi and A. Abdelhamed and Abdullah Abuolaim and Abhijith Punnappurath and M. S. Brown},
  journal={IEEE Transactions on Pattern Analysis and Machine Intelligence},
  year={2020},
  volume={44},
  pages={4688-4700},
}

@article{woo2017improving,
  title={Improving color constancy in an ambient light environment using the phong reflection model},
  author={Woo, Sung-Min and Lee, Sang-Ho and Yoo, Jun-Sang and Kim, Jong-Ok},
  journal={IEEE Transactions on Image Processing},
  volume={27},
  number={4},
  pages={1862--1877},
  year={2017},
  publisher={IEEE}
}

@article{Akbarinia2018ColourCB,
  title={Colour Constancy Beyond the Classical Receptive Field},
  author={Arash Akbarinia and C. Alejandro P{\'a}rraga},
  journal={IEEE Transactions on Pattern Analysis and Machine Intelligence},
  year={2018},
  volume={40},
  pages={2081-2094},
  url={https://api.semanticscholar.org/CorpusID:42979022}
}

@inproceedings{Gao2014EfficientCC,
  title={Efficient Color Constancy with Local Surface Reflectance Statistics},
  author={Shaobing Gao and Wangwang Han and Kaifu Yang and Chaoyi Li and Y. Li},
  booktitle={European Conference on Computer Vision},
  year={2014},
  url={https://api.semanticscholar.org/CorpusID:14518587}
}

@inproceedings{afifi2019sensor,
  title={Sensor-independent illumination estimation for {DNN} models},
  author={Afifi, Mahmoud and Brown, Michael S},
  booktitle={BMVC},
  year={2019}
}

@inproceedings{hu2017fc4,
  title={{FC}4: {F}ully convolutional color constancy with confidence-weighted pooling},
  author={Hu, Yuanming and Wang, Baoyuan and Lin, Stephen},
  booktitle={CVPR},
  year={2017}
}

@inproceedings{lo2021clcc,
  title={{CLCC}: {C}ontrastive learning for color constancy},
  author={Lo, Yi-Chen and Chang, Chia-Che and Chiu, Hsuan-Chao and Huang, Yu-Hao and Chen, Chia-Ping and Chang, Yu-Lin and Jou, Kevin},
  booktitle={CVPR},
  year={2021}
}

@inproceedings{afifi2025optimizing,
  title={Optimizing Illuminant Estimation in Dual-Exposure {HDR} Imaging},
  author={Afifi, Mahmoud and Hu, Zhenhua and Liang, Liang},
  booktitle={ECCV},
  year={2025},
}

@inproceedings{tang2022transfer,
  title={Transfer learning for color constancy via statistic perspective},
  author={Tang, Yuxiang and Kang, Xuejing and Li, Chunxiao and Lin, Zhaowen and Ming, Anlong},
  booktitle={AAAI},
  year={2022}
}

@inproceedings{xu2020end,
  title={End-to-end illuminant estimation based on deep metric learning},
  author={Xu, Bolei and Liu, Jingxin and Hou, Xianxu and Liu, Bozhi and Qiu, Guoping},
  booktitle={CVPR},
  year={2020}
}

@inproceedings{qian2017recurrent,
  title={Recurrent color constancy},
  author={Qian, Yanlin and Chen, Ke and Nikkanen, Jarno and Kamarainen, Joni-Kristian and Matas, Jiri},
  booktitle={ICCV},
  year={2017}
}

@inproceedings{yu2020cascading,
  title={Cascading convolutional color constancy},
  author={Yu, Huanglin and Chen, Ke and Wang, Kaiqi and Qian, Yanlin and Zhang, Zhaoxiang and Jia, Kui},
  booktitle={AAAI},
  year={2020}
}

@inproceedings{li2024nightcc,
  title={Night{CC}: {N}ighttime Color Constancy via Adaptive Channel Masking},
  author={Li, Shuwei and Tan, Robby T},
  booktitle={CVPRW},
  year={2024}
}

@article{yue2023color,
  title={Color constancy from a pure color view},
  author={Yue, Shuwei and Wei, Minchen},
  journal={JOSA A},
  volume={40},
  number={3},
  pages={602--610},
  year={2023},
}

@inproceedings{afifi2019color,
  title={When color constancy goes wrong: {C}orrecting improperly white-balanced images},
  author={Afifi, Mahmoud and Price, Brian and Cohen, Scott and Brown, Michael S},
  booktitle={CVPR},
  pages={1535--1544},
  year={2019}
}

@inproceedings{kim2021large,
  title={Large scale multi-illuminant (lsmi) dataset for developing white balance algorithm under mixed illumination},
  author={Kim, Dongyoung and Kim, Jinwoo and Nam, Seonghyeon and Lee, Dongwoo and Lee, Yeonkyung and Kang, Nahyup and Lee, Hyong-Euk and Yoo, ByungIn and Han, Jae-Joon and Kim, Seon Joo},
  booktitle={Proceedings of the IEEE/CVF International Conference on Computer Vision},
  pages={2410--2419},
  year={2021}
}

@inproceedings{kim2024attentive,
  title={Attentive Illumination Decomposition Model for Multi-Illuminant White Balancing},
  author={Kim, Dongyoung and Kim, Jinwoo and Yu, Junsang and Kim, Seon Joo},
  booktitle={Proceedings of the IEEE/CVF Conference on Computer Vision and Pattern Recognition (CVPR)},
  pages={25512--25521},
  year={2024}
}

@article{yue2024effective,
  title={Effective cross-sensor color constancy using a dual-mapping strategy},
  author={Yue, Shuwei and Wei, Minchen},
  journal={JOSA A},
  volume={41},
  number={2},
  pages={329--337},
  year={2024},
}

@inproceedings{hernandez2020multi,
  title={A multi-hypothesis approach to color constancy},
  author={Hernandez-Juarez, Daniel and Parisot, Sarah and Busam, Benjamin and Leonardis, Ales and Slabaugh, Gregory and McDonagh, Steven},
  booktitle={CVPR},
  year={2020}
}

@inproceedings{xiao2020multi,
  title={Multi-domain learning for accurate and few-shot color constancy},
  author={Xiao, Jin and Gu, Shuhang and Zhang, Lei},
  booktitle={CVPR},
  year={2020}
}

@article{mcdonagh2018formulating,
  title={Formulating camera-adaptive color constancy as a few-shot meta-learning problem},
  author={McDonagh, Steven and Parisot, Sarah and Zhou, Fengwei and Zhang, Xing and Leonardis, Ales and Li, Zhenguo and Slabaugh, Gregory},
  journal={arXiv preprint arXiv:1811.11788},
  year={2018}
}

@inproceedings{bianco2019quasi,
  title={Quasi-unsupervised color constancy},
  author={Bianco, Simone and Cusano, Claudio},
  booktitle={CVPR},
  year={2019}
}

@inproceedings{karaimer2018improving,
  title={Improving color reproduction accuracy on cameras},
  author={Karaimer, Hakki Can and Brown, Michael S},
  booktitle={CVPR},
  year={2018}
}

@inproceedings{mcelvain2013camera,
  title={Camera color correction using two-dimensional transforms},
  author={McElvain, Jon S and Gish, Walter},
  booktitle={Color and Imaging Conference},
  year={2013},
}

@article{adobe2023digital,
  title={Digital Negative ({DNG}) Specification},
  author={Adobe Systems Incorporated},
  year={2023}
}

@inproceedings{florin2009color,
  title={Color processing in a digital camera pipeline},
  author={Florin, Toadere},
  booktitle={Advanced Topics in Optoelectronics, Microelectronics, and Nanotechnologies IV},
  volume={7297},
  pages={223--227},
  year={2009},
}

@article{bianco2013color,
  title={Color correction pipeline optimization for digital cameras},
  author={Bianco, Simone and Bruna, Arcangelo R and Naccari, Filippo and Schettini, Raimondo},
  journal={Journal of Electronic Imaging},
  volume={22},
  number={2},
  pages={023014--023014},
  year={2013},
}

@inproceedings{karaimer2016software,
  title={A software platform for manipulating the camera imaging pipeline},
  author={Karaimer, Hakki Can and Brown, Michael S},
  booktitle={ECCV},
  year={2016},
}

@article{brown2023color,
  title={Color processing for digital cameras},
  author={Brown, MichaelS},
  journal={Fundamentals and Applications of Colour Engineering},
  pages={81--98},
  year={2023},
}

@article{delbracio2021mobile,
  title={Mobile computational photography: {A} tour},
  author={Delbracio, Mauricio and Kelly, Damien and Brown, Michael S and Milanfar, Peyman},
  journal={Annual review of vision science},
  volume={7},
  number={1},
  pages={571--604},
  year={2021},
}

@article{hasinoff2016burst,
  title={Burst photography for high dynamic range and low-light imaging on mobile cameras},
  author={Hasinoff, Samuel W and Sharlet, Dillon and Geiss, Ryan and Adams, Andrew and Barron, Jonathan T and Kainz, Florian and Chen, Jiawen and Levoy, Marc},
  journal={ACM Transactions on Graphics (ToG)},
  volume={35},
  number={6},
  pages={1--12},
  year={2016},
}

@article{gijsenij2011computational,
  title={Computational color constancy: {S}urvey and experiments},
  author={Gijsenij, Arjan and Gevers, Theo and Van De Weijer, Joost},
  journal={IEEE Transactions on Image Processing},
  volume={20},
  number={9},
  pages={2475--2489},
  year={2011},
}

@inproceedings{nguyen2014raw,
  title={Raw-to-raw: Mapping between image sensor color responses},
  author={Nguyen, Rang and Prasad, Dilip K and Brown, Michael S},
  booktitle={CVPR},
  year={2014}
}

@inproceedings{afifi2021semi,
  title={Semi-supervised raw-to-raw mapping},
  author={Afifi, Mahmoud and Abuolaim, Abdullah},
  booktitle={BMVC},
  year={2021}
}

@inproceedings{cheng2015beyond,
  title={Beyond white: {G}round truth colors for color constancy correction},
  author={Cheng, Dongliang and Price, Brian and Cohen, Scott and Brown, Michael S},
  booktitle={CVPR},
  year={2015}
}

@inproceedings{barnard1995computational,
  title={COMPUTATIONAL COLOR CONSTANCY: TAKING THEORY INTO PRACTICE},
  author={Kobus Barnard},
  year={1995},
  url={https://api.semanticscholar.org/CorpusID:62363219}
}

@article{barnard2002comparison,
  title={A Comparison of Computational Color Constancy
Algorithms--Part {II}: {E}xperiments With
Image Data},
  author={Barnard, Kobus and Martin, Lindsay and Coath, Adam and Funt, Brian},
  journal={IEEE Transactions on Image Processing},
  volume={11},
  number={9},
  pages={985--996},
  year={2002},
}

@article{finlayson2015color,
  title={Color correction using root-polynomial regression},
  author={Finlayson, Graham D and Mackiewicz, Michal and Hurlbert, Anya},
  journal={IEEE Transactions on Image Processing},
  volume={24},
  number={5},
  pages={1460--1470},
  year={2015},
}

@article{hung1993colorimetric,
  title={Colorimetric calibration in electronic imaging devices using a look-up-table model and interpolations},
  author={Hung, Po-Chieh},
  journal={Journal of Electronic imaging},
  volume={2},
  number={1},
  pages={53--61},
  year={1993},
}

@article{hong2001study,
  title={A study of digital camera colorimetric characterization based on polynomial modeling},
  author={Hong, Guowei and Luo, M Ronnier and Rhodes, Peter A},
  journal={Color Research \& Application},
  volume={26},
  number={1},
  pages={76--84},
  year={2001},
}

@article{finlayson2020designing,
  title={Designing color filters that make cameras more colorimetric},
  author={Finlayson, Graham D and Zhu, Yuteng},
  journal={IEEE Transactions on Image Processing},
  volume={30},
  pages={853--867},
  year={2020},
}
}

\end{document}